\documentclass[runningheads]{llncs}
\usepackage{graphicx}
\usepackage{caption}
\usepackage{subcaption}
\usepackage{multirow}

\usepackage{tikz}
\usepackage{comment}
\usepackage{amsmath,amssymb} 
\usepackage{color}
\usepackage{xcolor}
\usepackage[]{color-edits}
\usepackage[accsupp]{axessibility}  

\begin{document}
\pagestyle{headings}
\mainmatter

\title{Object Detection for Autonomous Dozers} 

\titlerunning{Object Detection for Autonomous Dozers}
%
\author{Chun-Hao Liu\thanks{corresponding author} \and Burhaneddin Yaman}
\authorrunning{Chun-Hao and Burhan}
%
\institute{Bosch Center for Artificial Intelligence, Sunnyvale, USA
\email{\{Chun-Hao.Liu,Burhaneddin.Yaman\}@us.bosch.com}}
\maketitle

\begin{abstract}
We introduce a new type of autonomous vehicle -- an autonomous dozer that is expected to complete construction site tasks in an efficient, robust, and safe manner. To better handle the path planning for the dozer and ensure construction site safety, object detection plays one of the most critical components among perception tasks. In this work, we first collect the construction site data by driving around our dozers. Then we analyze the data thoroughly to understand its distribution. Finally, two well-known object detection models are trained, and their performances are benchmarked with a wide range of training strategies and hyperparameters.

\end{abstract}

\section{Introduction}
The trade-off for accuracy and speed is one of the critical challenges for perception tasks in autonomous vehicles (AVs)~\cite{8099834}. Among multiple perception tasks~\cite{9684905}, object detection (OD) is one of the fundamental tasks that is required for most AVs~\cite{Hu_2022_CVPR}. AV systems has to ensure high accuracy for OD in order to ensure its detection results can be confidently shared with other following tasks, such as path planning and safety critical functions. There are several challenges and requirements for OD models. Firstly, OD models must work robustly, e.g., under various environmental conditions such as adverse weather~\cite{Bijelic_2020_STF} and low light at night~\cite{Wu_2021_CVPR}. Secondly, it needs to detect a wide variety of on/off road objects (e.g., person, vehicle, static-objects, etc.) with diverse sizes~\cite{9684905}. Finally, inference speed is a stringent requirement for OD models~\cite{8014794} in order to meet safety criteria. Different AVs have their own vehicle computing platform and the corresponding latency constraint. OD has to satisfy these hardware constraints while not sacrificing detection accuracy much.

We introduce a new type of AV, an autonomous dozer working on a construction site. The autonomous dozer is expected to flatten out the sand piles stacked in front of it in an efficient, robust, and safe manner. To achieve this goal, we first collect real-world construction site data by driving around the dozer and finish the flattening tasks. The driving task is completed by human experts. Then we develop our deep learning based algorithms learned by the recorded data for both path planning and perception tasks. In this work, we focus on studying the performance of OD with different popular models trained on the collected construction site data. To the best of our knowledge, this is the first work to benchmark OD performance on autonomous dozers with thorough inspection on all key factors that can affect detection performance. Our main contributions are listed as follows.
\begin{enumerate}
    \item A new type of AV is introduced. The corresponding construction dataset is collected and analyzed rigorously.
    \item Two common OD models are introduced and trained on this dataset.
    \item The OD models performance covering a wide range of key hyperparameters and popular training strategies are benchmarked.
\end{enumerate}

\section{Construction Site Data}
In the following sections, we first introduce our sensor setup for the autonomous dozer. Then we introduce the data collection process, label definitions, and the labeling strategy. Finally, we present the data statistics to have a clear view of the construction site data.

\subsection{Sensor Setup}
\begin{figure}
\vspace{-\intextsep}
\centering
\includegraphics[clip,keepaspectratio, width=0.5\textwidth]{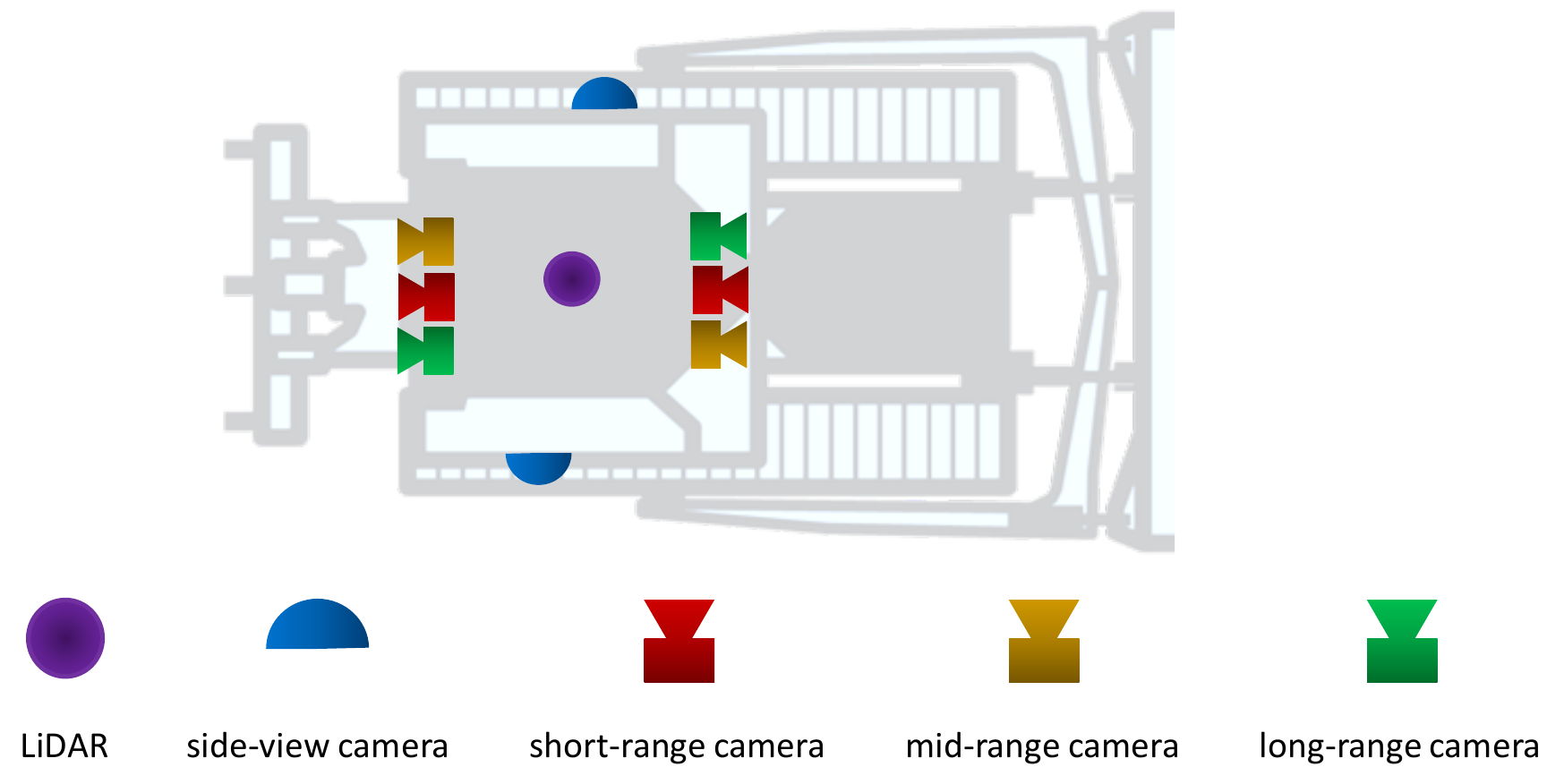}
\vspace{-0.3cm}
\caption{Sensor setup for the autonomous dozer.}
\vspace{-1cm}
\label{fig:sensor_setup}
\end{figure}

Fig.~\ref{fig:sensor_setup} shows the sensor setup for the dozer. It is used for data collection provided to OD only, not including other modules. Four types of cameras are being setup, including long-range, middle-range, short-range, and fisheye cameras. Different kinds of camera can focus on objects within different target distances in order to achieve the best performance. For example, short range-camera can take care of nearby objects to ensure safety critical commands. Mid-range and long-range cameras can provide useful guidance to both short-term and long-term path planning algorithms. Side-view cameras are fisheye cameras which have wider field of view so that it can help to understand the surrounding to avoid potential collisions. Together with front-view, rear-view, and side-view cameras, the dozer is able to observe its $360^\circ$ surrounding and hence safety is guaranteed with higher confidence. To further enhance 3D environmental understanding, we also setup two omnidirectional LiDARs (Light Detection and Ranging) on the dozer. The 3D information from LiDARs can not only help to estimate the object distance from the dozer, but can also increase the OD performance through smart sensor fusion.
\subsection{Collection Strategy}
We select two construction sites to collect data, named as site $\mathcal{N}$ and site $\mathcal{Y}$. Site $\mathcal{N}$ is a construction site without constraint, where data collection happens simultaneously with the ongoing real construction works. In contrast, site $\mathcal{Y}$ is a construction site with constraints, where we place obstacles and commonly seen vehicles on construction sites around our dozer. Each object is placed with some patterns, e.g., a fixed orientation and distance. The advantage for site $\mathcal{N}$ is that we can collect more diverse type of objects, but it is hard to catch different perspectives for different objects due to dynamic moving objects and safety concerns. On the other hand, it is easy to overcome the drawback of perspectives on site $\mathcal{Y}$ since all objects are static and under controlled. However, site $\mathcal{Y}$ lacks diversity as only limited type of objects can be placed.

When we start to collect data, we turn on the video and point cloud recording for all cameras and LiDARs, respectively. Each recording is around $5$ minutes, and we store all the collected sensor data in a highly compressed data format for easily uploading to the cloud and post-processing purposes. We drive our dozer randomly which covers the entire construction area on site $\mathcal{N}$. Pre-defined routes to capture different object perspectives, object types, and geographic orientations are implemented as the driving strategy on site $\mathcal{Y}$. Time is also being considered in our data collection process. Specifically, data is collected during three different times -- morning, noon, and dusk. We collect around $90$ and $30$ recordings from site $\mathcal{N}$ and site $\mathcal{Y}$, respectively. The entire data collection is done in three months during summer $2021$.

\subsection{Data Labeling}
We define a broad range of classes by consulting with the construction experts. These classes of objects are common ones seen and used on construction sites, and they have higher detection priority to satisfy the critical safety concerns. Table~\ref{tab:class_def} shows our definition for the labeled classes. We firstly define three high-level classes, and then multiple low-level classes under each high-level class are elaborated. For each low-level class, we provide several commonly used attributes, i.e., difficulty, occlusion, and truncation, similar to the ones used in publicly available autonomous driving datasets~\cite{Geiger2012CVPR,Cordts2016Cityscapes,bdd100k}. Note for ``Person'', we label the color of their safety hats and vests. For ``Dump-truck'', we label if it has sand piles on its trunk. For the ``Vehicle'', we label if there is ``Person'' sitting inside or not. Among all labeled classes, we further define $13$ training classes used for the OD task, namely {\bf Person}, {\bf Dump-truck}, {\bf Compactor}, {\bf Excavators}, {\bf Dozer}, {\bf Sprinkler}, {\bf Vehicle}, {\bf Pipes}, {\bf Color-cones}, {\bf Bar}, {\bf Drum-cans}, {\bf Generator}, and {\bf Survey-equipment}. Specifically, training class {\bf Vehicle} is composed of low-level classes ``Car'', ``Van'', and ``Other''.

For labeling process, we follow a standard procedure for all the images. All images are distributed across $9$ labelers to finish the labeling task. After the labeling is finished, the labeled files will be passed to the first stage quality checker. The quality checker will check all labels annotated on the images as well as the missing labels. If they pass this check, these labels will be passed to the second stage quality checker. If they fail it, these labels will go back to the original labelers to fix all issues until they pass the check. At last, the second stage quality checker runs the same procedure as the first stage. However, they only sample partial of the provided images and labels to check. 
\begin{table}
\centering
\setlength\tabcolsep{0pt}
\begin{tabular}{c|c}
\hline
High-Level Class & Low-Level Class\\
\hline
\hline
Person & Person \\
Vehicle &  Car, Dump-truck, Compactor, Van, Excavators, Dozer, Sprinkler, Other \\
Static-objects & Drum-cans, Generator, Pipes, Color-cones, Survey-equipment, Bar, Other \\
\hline
\end{tabular}
\caption{Definition of labeled classes.}
\vspace{-1cm}
\label{tab:class_def}
\end{table}
\subsection{Data Analysis}
\begin{figure}
\vspace{-\intextsep}
\begin{subfigure}[t]{0.32\linewidth}
    \includegraphics[width=\linewidth]{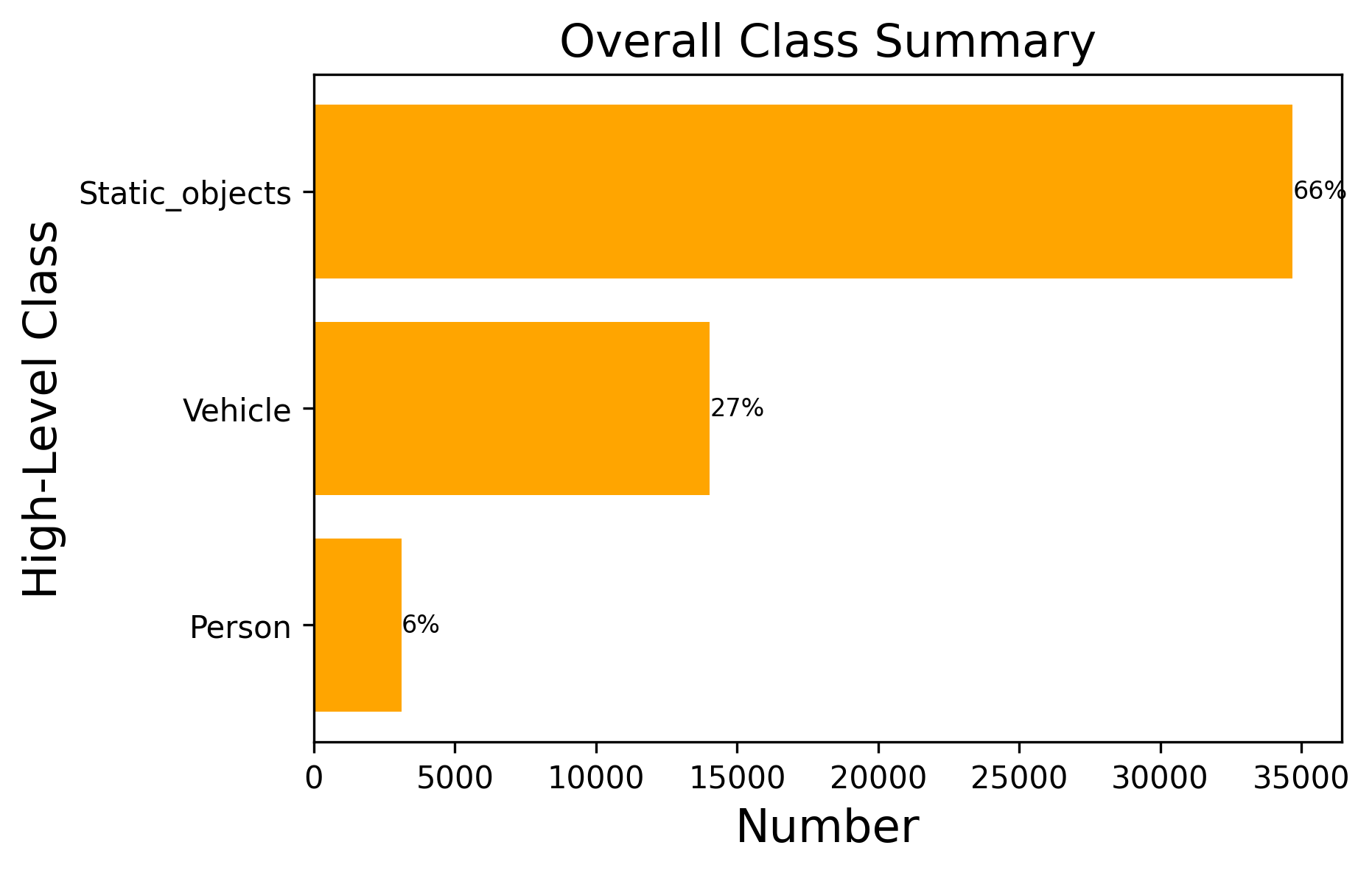}
\end{subfigure}%
\begin{subfigure}[t]{0.32\linewidth}
    \includegraphics[width=\linewidth]{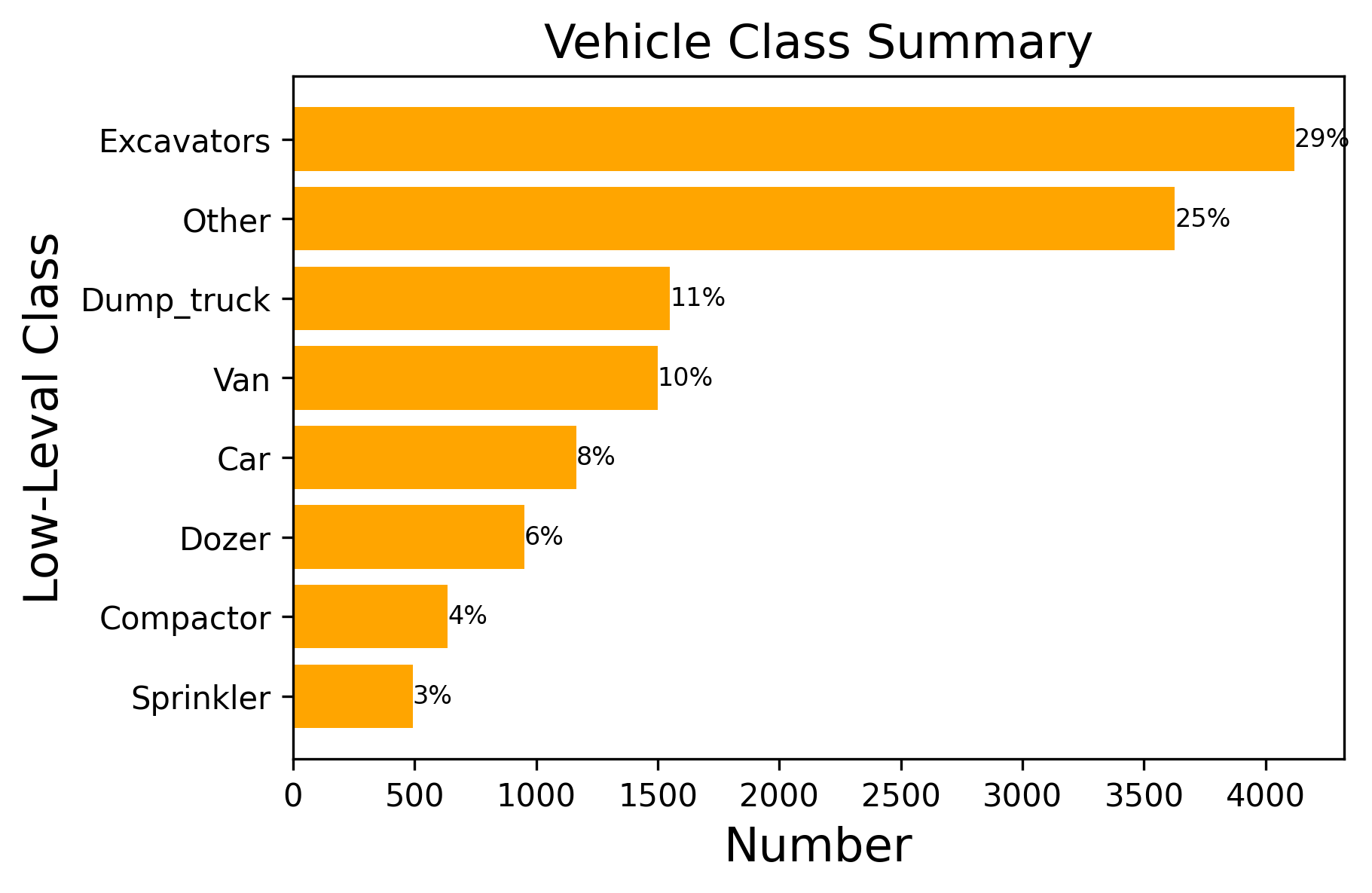}
\end{subfigure}
\begin{subfigure}[t]{0.32\linewidth}
    \includegraphics[width=\linewidth]{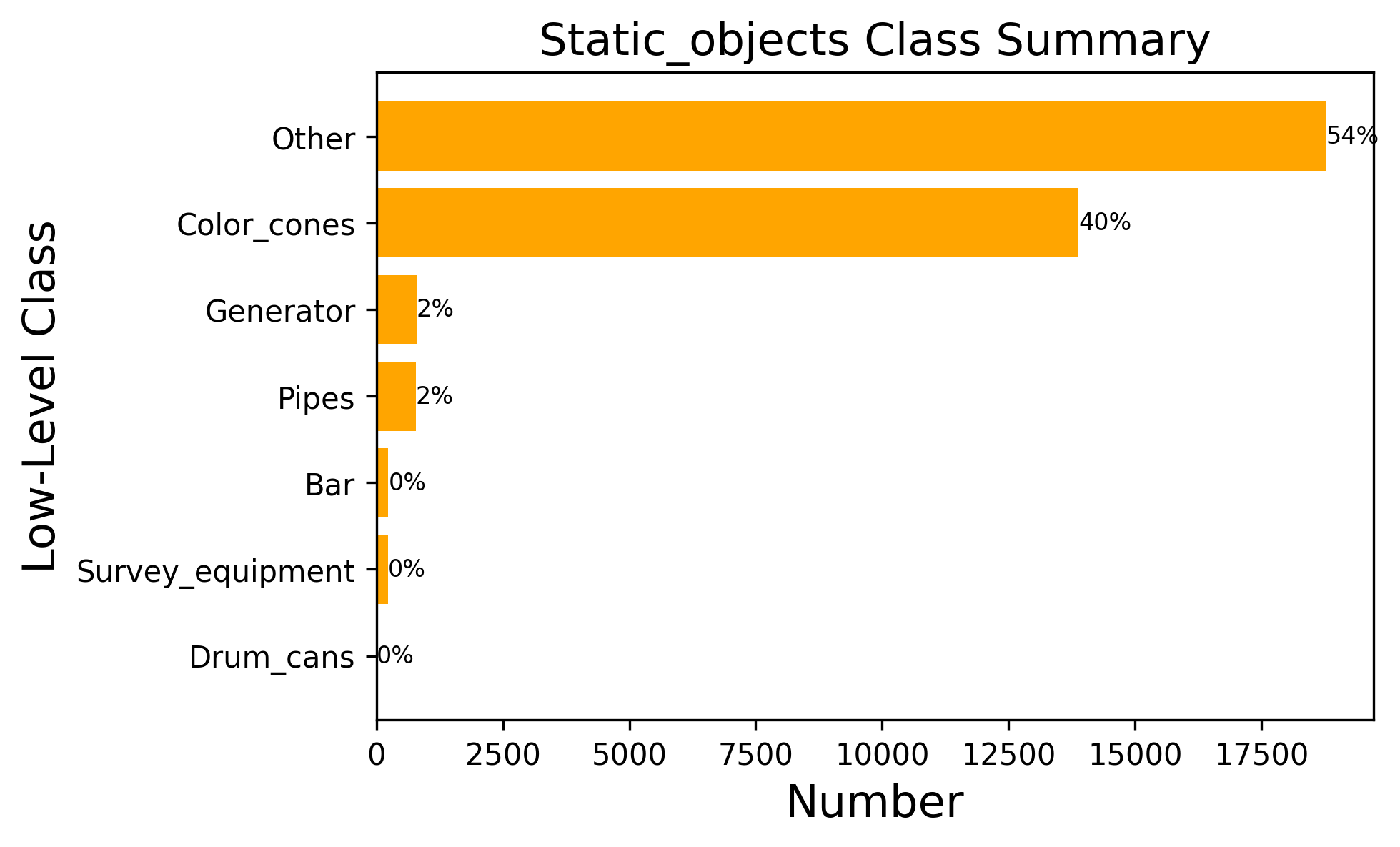}
\end{subfigure}
\caption{Number of high-level and low-level classes.
}
\vspace{-\intextsep}
\label{fig:number_of_classes}
\end{figure}

In total, we extract around $3000$ images for each camera recorded on site $\mathcal{N}$, and $700$ images on site $\mathcal{Y}$. Note each image is subsampled from the video with constant time interval covering the entire recording duration. We analyze the data statistics of front-view middle-range camera\footnote{The data statistics is similar in other cameras.} on site $\mathcal{N}$ as an example. From Fig.~\ref{fig:number_of_classes}, we can see that ``Static-objects" dominate $66\%$ in high-level classes, and we only have $6\%$ of ``Person". Among different types of vehicles, we find ``Excavators" are the most common one on the construction site. The distribution of ``Static-objects" is the most imbalanced one. ``Other" and ``Color-cones" together dominate $94\%$ of it. For the rest of the ``Static-objects", we suffer from the severe data imbalance problem. For example, we only have $3$ instances for ``Drum-cans".

\begin{figure}
\vspace{-\intextsep}
\begin{subfigure}[t]{0.32\linewidth}
    \includegraphics[width=\linewidth]{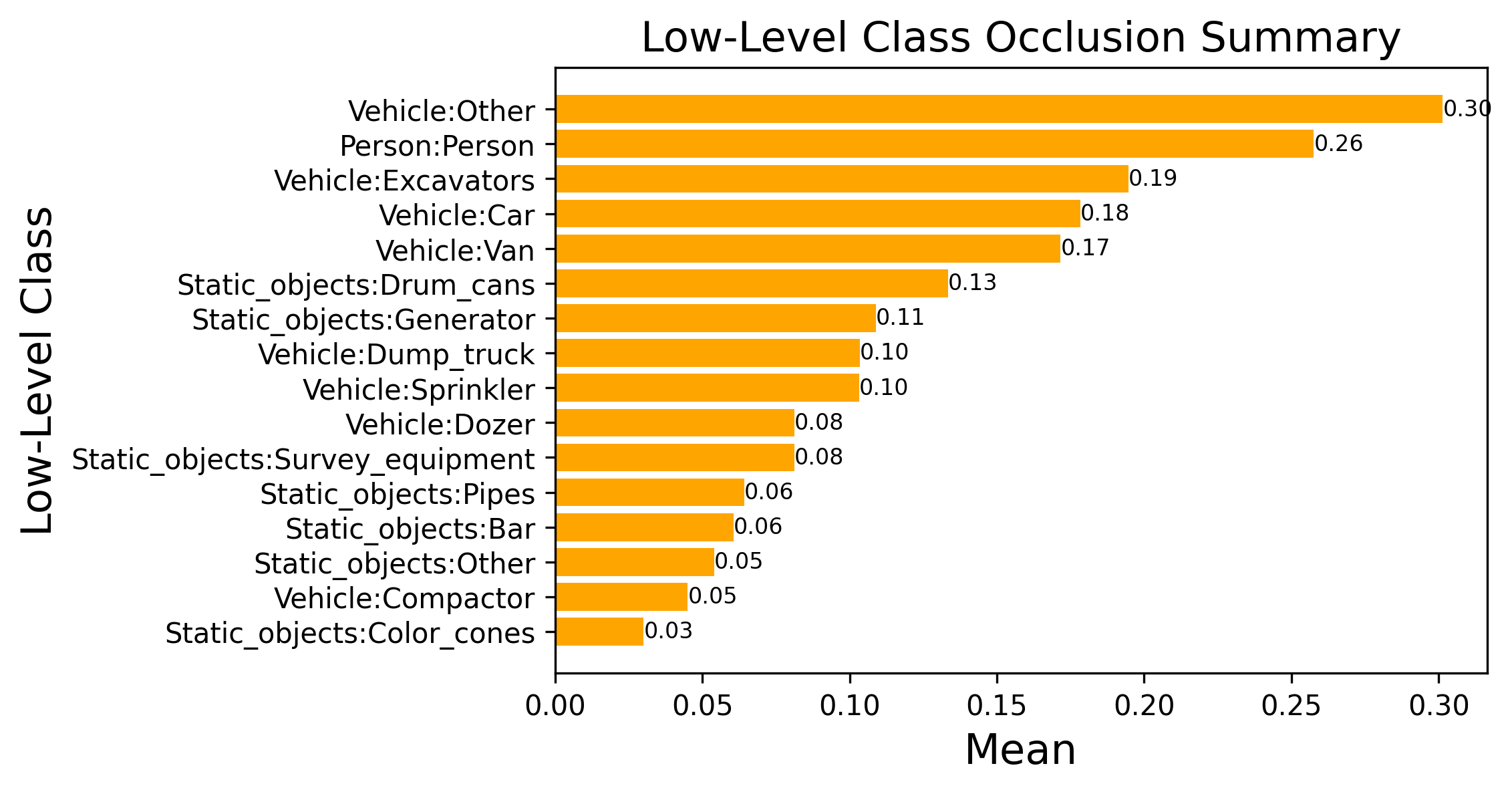}
\end{subfigure}%
\begin{subfigure}[t]{0.32\linewidth}
    \includegraphics[width=\linewidth]{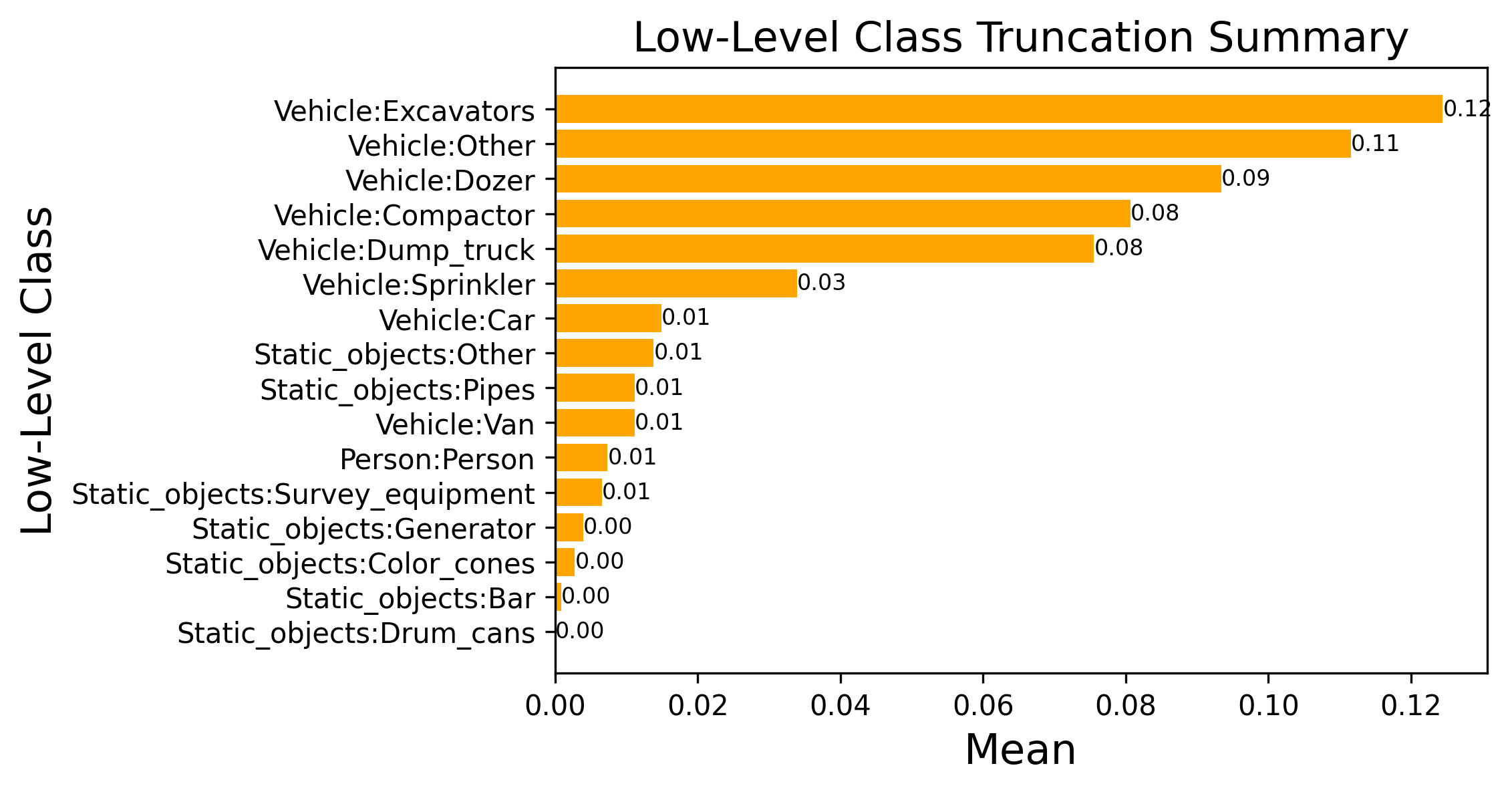}
\end{subfigure}
\begin{subfigure}[t]{0.32\linewidth}
    \includegraphics[width=\linewidth]{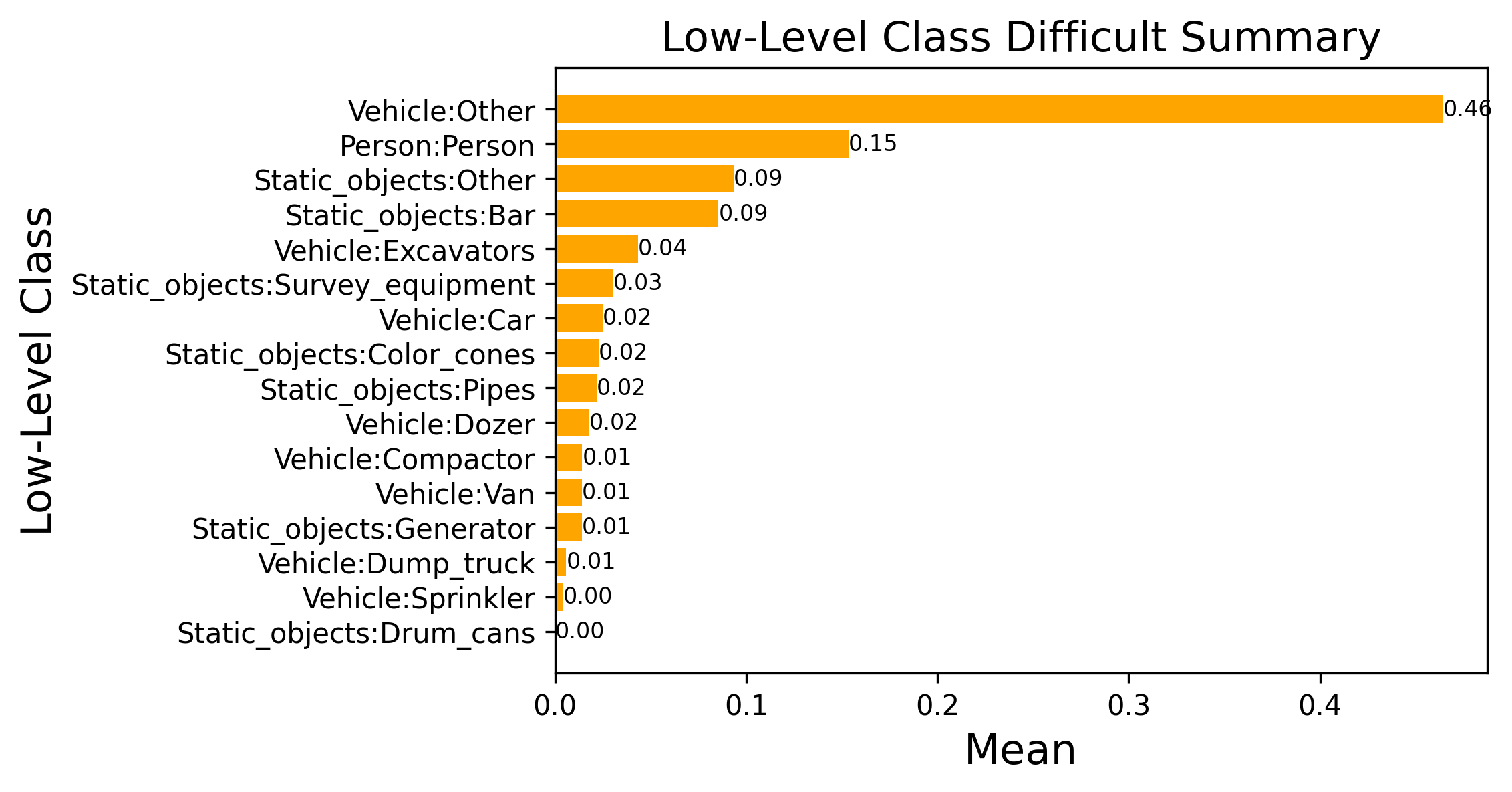}
\end{subfigure}
\caption{Attributes for low-level classes.
}
\vspace{-\intextsep}
\label{fig:attributes}
\end{figure}

Next, we look at the attributes of occlusion, truncation, and difficulty for each low-level class. Fig.~\ref{fig:attributes} shows the average severity value for each attribute. They are labeled in the range of $[0, 1]$.  The two classes that are being occluded largely are ``Other" from ``Vehicle" and ``Person". It can be explained by that ``Other from "Vehicle" is usually a vehicle with an unknown type. However, since objects shown in construction site are highly controlled, this will mostly happen when the labelers are not sure about faraway vehicles. This kind of vehicle can be easily occluded as they are usually small on the images. The same reason can be applied to ``Person". For truncation, we can see all classes are with around $10\%$ or less on average. This implies that truncation is not a serious problem in identifying objects. Lastly, ``Other" from ``Vehicle" shows a high difficulty ($46\%$) to be seen by labelers. We attribute it to the same reason as the occlusion issue.

\begin{figure}
\vspace{-\intextsep}
\centering
\includegraphics[clip,keepaspectratio, width=0.5\textwidth]{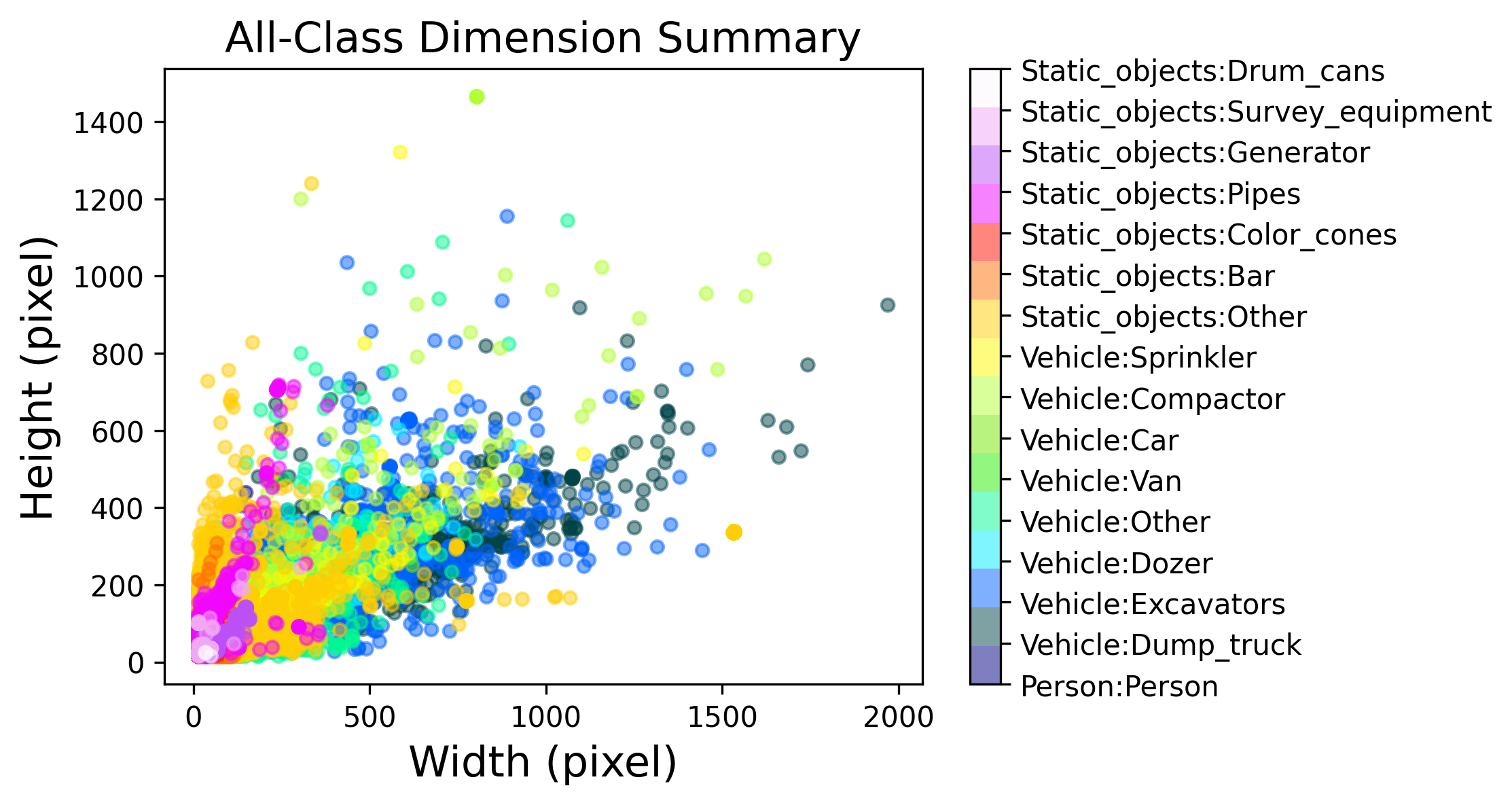}
\caption{Dimension distribution for low-level classes}
\vspace{-\intextsep}
\label{fig:dimension}
\end{figure}

Fig.~\ref{fig:dimension} shows the scatter plot of object height and width for all low-level classes. We can indicate that ``Person" and ``Static-objects" are much smaller than ``Vehicle", but they are with higher density. The mean for object height and width are $61.11$ and $71.92$ pixels, respectively. The $33\%$, $50\%$, and $75\%$ quantiles for object height and width are $(27, 15)$, $(37, 21)$, and $(52, 40)$, respectively. From the statistics we know that we have much fewer large objects compared to small objects.

\begin{figure}
\vspace{-\intextsep}
\centering
\includegraphics[clip,keepaspectratio, width=0.5\textwidth]{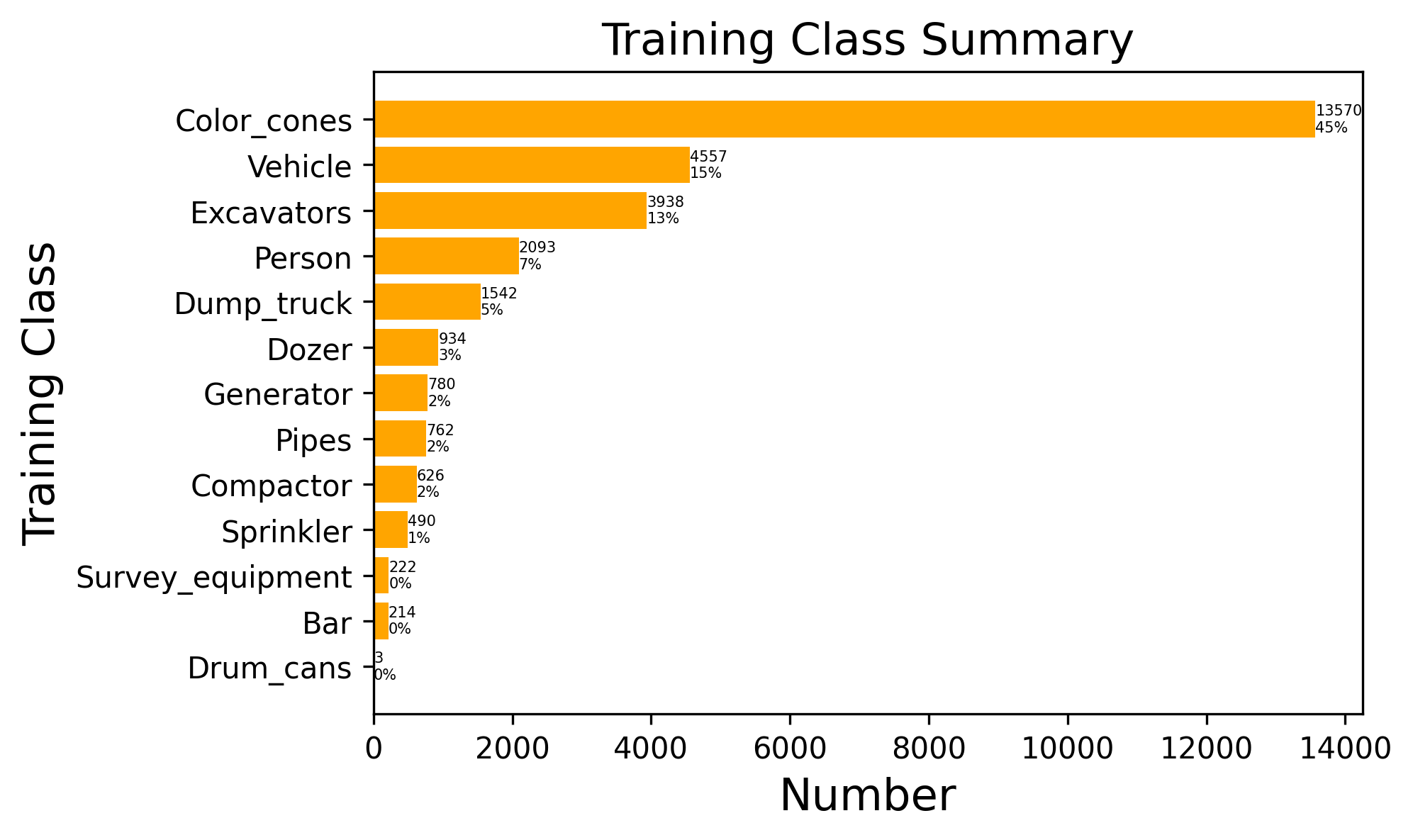}
\caption{The distribution of training classes.}
\vspace{-\intextsep}
\label{fig:training_classes}
\end{figure}

We look at the distribution of the defined $13$ training classes. Fig.~\ref{fig:training_classes} shows that the number of training classes follows an exponentially decaying distribution. ``Color-cones" is the dominant class with $45\%$, whereas  all other ``Static-objects" are all under $2\%$. We have reasonable amount of typical ``Vehicle" as $15\%$. However, for large construction type of vehicles such as ``Dozer" and ``Compactor" are rarely seen, except ``Excavators". In summary, we can envision that the challenge for the OD will be of highly imbalanced dataset, and a large number of tiny objects.

\begin{figure}
\begin{subfigure}[t]{0.32\linewidth}
    \includegraphics[width=\linewidth]{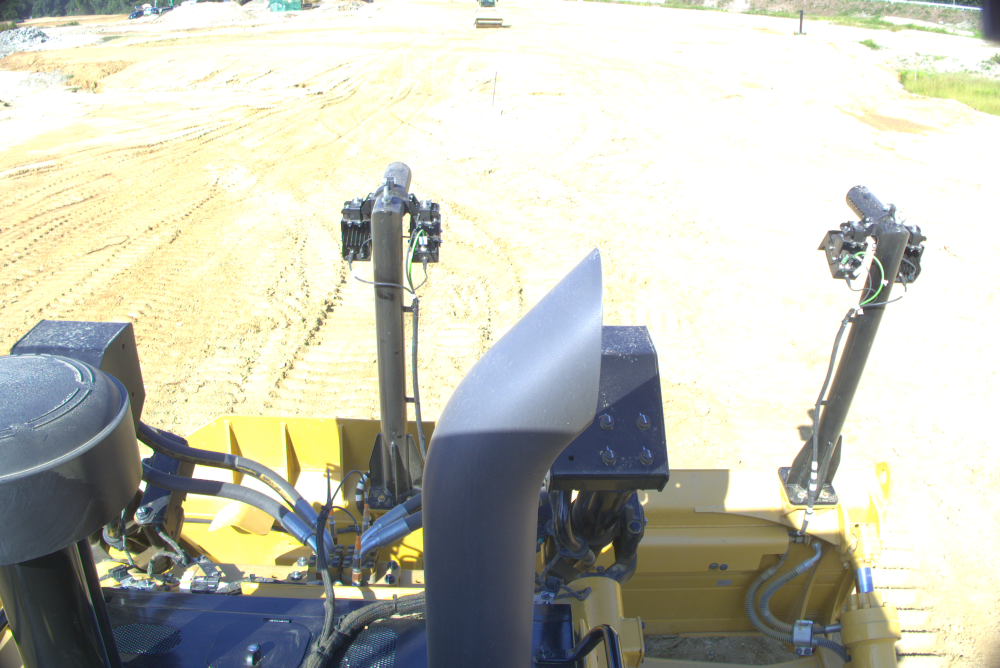}
    \caption{Strong light reflection.}
\label{fig:strong_light_examples}
\end{subfigure}\hspace{0.1cm}
\begin{subfigure}[t]{0.32\linewidth}
    \includegraphics[width=\linewidth]{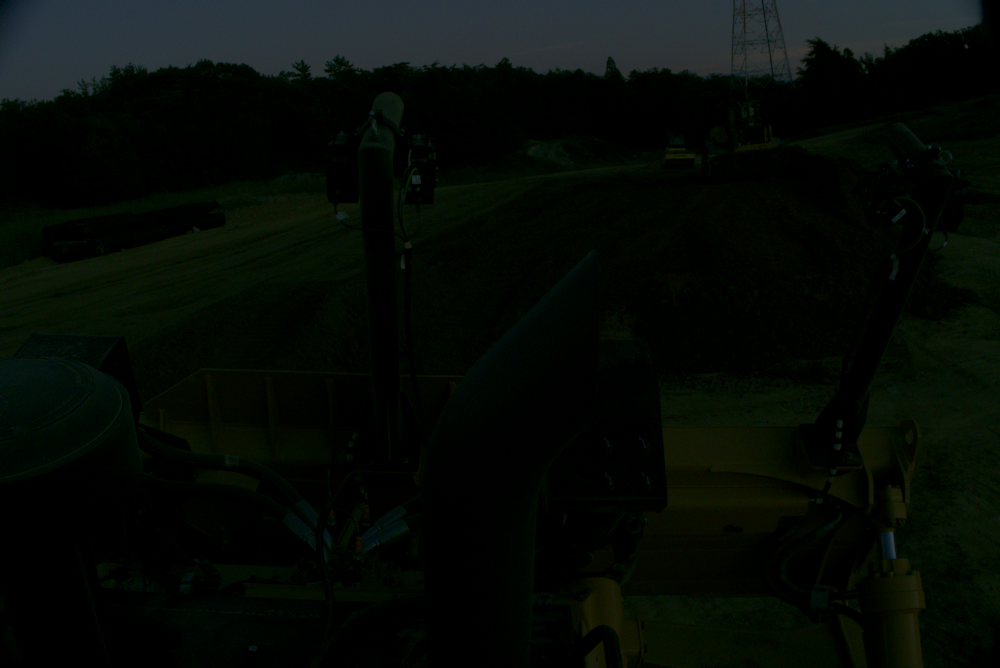}
    \caption{Low light at dusk.}
\label{fig:low_light_examples}
\end{subfigure}\hspace{0.1cm}
\begin{subfigure}[t]{0.32\linewidth}
    \includegraphics[width=\linewidth]{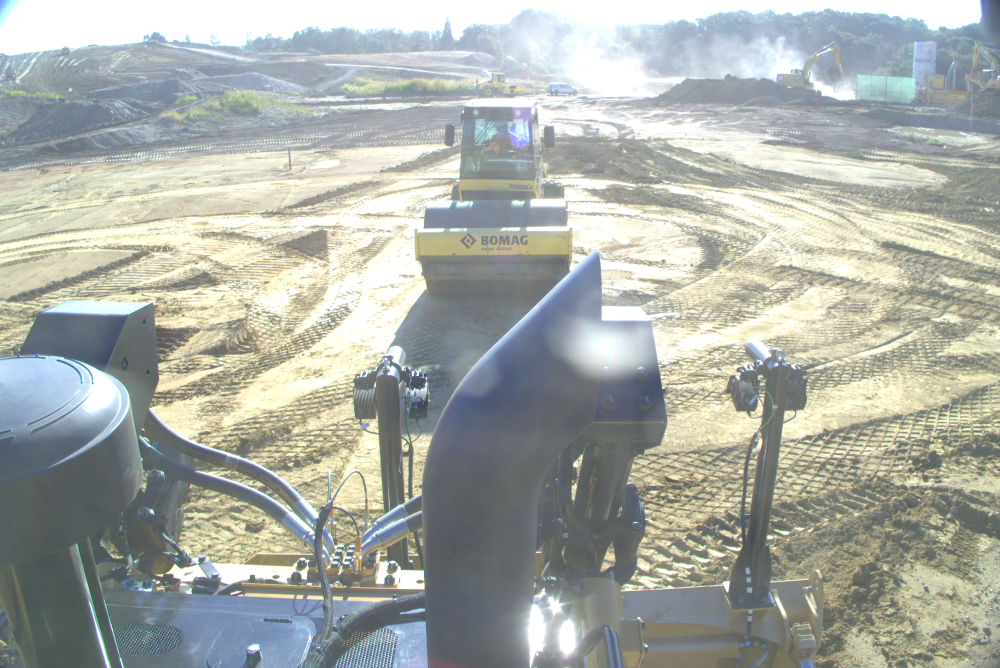}
    \caption{Dusty environment.}
\label{fig:dusty_examples}
\end{subfigure}
\caption{Challenging images taken from front-view middle-range camera.
}
\vspace{-\intextsep}
\label{fig:examples}
\end{figure}

Finally, we present some challenging visual examples collected by the dozer. Fig.~\ref{fig:strong_light_examples} shows a typical image that is collected on construction site. Usually the sun reflection is strong, and the strong exposure will lead to difficulties in recognition of objects. On the other hand, the construction site can be dark when it is near the sunset time, as shown in Fig.~\ref{fig:low_light_examples}. Lastly, dusty air is commonly seen in construction areas due to movement of vehicles such as excavators and dump-trucks. The dusty effect makes the captured images noisy and thus reduce visibility. This phenomenon is shown in Fig.~\ref{fig:dusty_examples}. There are still other possible challenges from the camera images, such as the big occlusion from ego-vehicle due to planted sticks equipped with other sensors.
\section{Object Detection Models}
OD models can be categorized into two types: anchor-based~\cite{7410526,7780460} and anchor-free approaches~\cite{Law_2018_ECCV,Duan_2019_ICCV}. Anchor-based approaches match the predicted bounding boxes with ground truth bounding boxes using a set of pre-defined anchors. Then the model classifies the prediction and rectifies the offset for the box coordinates through regression. Anchor-free approaches directly predict a set of key points for each object, without the need of prior anchors. The predicted key points are then grouped to determine the final bounding boxes. In addition to anchors, there are also two mainstream designs for the OD model, namely single-stage~\cite{7780460,8417976,10.1007/978-3-319-46448-0_2,9156454} and two-stage models~\cite{7410526,10.5555/2969239.2969250,8237584,10.1007/978-3-030-58555-6_16}. Single-stage models predict bounding boxes directly with one network. They are well-known for their fast inference speed and therefore suitable for real-time applications. On the other hand, two-stage models first predict bounding box proposals, and then perform box regression and classification on top of them. Conventionally, two-stage models can generate more accurate proposals, and thus has higher accuracy but lower speed compared to single-stage models. However, in recent years, single-stage models have been significantly developed further, attaining comparable or even higher accuracy than two-stage models. In this work we compare two detection models that are commonly used.

{\bf Scaled-YOLOv4.} The first one is scaled-YOLOv4~\cite{Wang_2021_CVPR}. YOLO~\cite{7780460} is a popular series of one-stage detectors that is well-known for their fast inference speed. The following generations such as YOLOv2~\cite{8100173}, YOLOv3~\cite{redmon2018yolov3}, YOLOv4~\cite{bochkovskiy2020yolov4} etc., are evolved based on similar model structures. Scaled-YOLOv4 improves on top of YOLOv4, by fully leveraging the power of the Cross-Stage-Partial-connections (CSP) approach to significantly reduce computations. The CSP scheme not only applies to the backbone network, but also applies to its neck and path aggregation networks. Furthermore, the authors studied the scaling factors on all aspects of the network so that they made YOLOv4 scalable to both larger and smaller models.

{\bf Faster R-CNN.} The second one is Faster R-CNN~\cite{10.5555/2969239.2969250}. Region-based Convolutional Neural Network (R-CNN) and its follow-up works -- Fast R-CNN and Faster R-CNN are the two-stage detectors that provide high localization and classification accuracy. In two-stage detectors, first stage performs object region proposals and second stage uses the features extracted from these proposed regions for object classification and bounding box regression. Despite the high accuracy, early R-CNN methods suffer from lengthy inference time which hinders their application in real-time. Faster R-CNN tackles this major issue and enables the usage of two-stage detectors in real-time by replacing the slow selective search region proposal framework used in earlier R-CNNs with a neural network based region proposal network (RPN).

\section{Experiments}

\subsection{Training Setup}
We adopt the images/labels taken from front middle-range camera from construction site $\mathcal{N}$ as our target data. We split $80\%$ of them as the training set, and $20\%$ of them as validation set. Our baseline training configuration is set to use batch size $16$, $300$ epochs, the Adam optimizer~\cite{DBLP:journals/corr/KingmaB14}, cosine learning rate decay with initial learning rate $10^{-3}$, weak augmentation by random horizontal flip, and $10$ epochs for warm up training from learning rate $10^{-6}$. The input image to the model is resized to $512\times512$ pixels while the ratio of original image height and width are kept the same. Scaled YOLOv4-p5, i.e., a three-head scaled YOLOv4, and Faster R-CNN with backbone ResNet-50 are selected as baseline models for one and two stage detectors, respectively.

\subsection{Computation Resources}
We implement the proposed algorithm with Python 3.7 and TensorFlow 2.4.0. An Intel(R) Xeon(R) Gold 6150 CPU @2.70GHz with 64GB RAM and 8 NVIDIA Tesla V100 GPUs with 32GB VRAM are used for training models.

\subsection{Performance Benchmark}
In this section we present the performance benchmarked results on our defined validation set, using MS COCO style evaluation metrics~\cite{10.1007/978-3-319-10602-1_48}\footnote{The metrics include mean average precision (mAP), average precision (AP) with intersection over union (IoU) $50\%$ (AP$_{50}$), AP with IoU $75\%$ (AP$_{75}$), mAP on small objects (mAPs), mAP on medium objects (mAPm), and mAP on large objects (mAPl).}.

{\bf Key hyperparameters.} We run extensive experiments and summarize the key performance hyperparameters in Table~\ref{tab:kpi}. The identified key hyperparameters are {\bf transfer learning}, {\bf optimizer}, {\bf label smoothing}, {\bf normalization layer}, and {\bf batch size}. For scaled-YOLOv4,  first, we can observe using transfer learning from a pre-trained model on COCO~\cite{10.1007/978-3-319-10602-1_48} dataset results in a large mAP improvement. We have also tested the transferred weights from other datasets such as KITTI~\cite{Geiger2012CVPR} and Cityscapes~\cite{Cordts2016Cityscapes}. However, they only provide either no or marginal improvement. This experiment gives us the hint that using pre-trained weights from a larger dataset may benefit more on the following transferred task. Second, soft labels are proven to be a strong regularization technique to prevent over-confident on model prediction~\cite{9464693}. The simple soft label generation scheme, i.e., label smoothing is first proposed in~\cite{7780677}. The authors assigned the label for each instance by averaging between their hard label and the uniform distribution over all classes. The label smoothness is controlled by a coefficient $\alpha$. Larger $\alpha$ represents more uncertainty is being assigned, and $\alpha=0$ means hard-label assignment. The $k$-th element for target label $y$ can be formulated as
\begin{align}
y_k=\left\{\begin{matrix}
 1-\alpha+\frac{\alpha}{C},& \mbox{if} \ k=c, \\ 
 \frac{\alpha}{C}, & \mbox{if} \ k\neq c,
\end{matrix}\right.
\label{eq:smoothing}
\end{align}
where $c$ is the index of target label and $C$ is total number of classes. We observe label smoothing can bring large improvement by choosing the correct smoothing factor $\alpha=0.2$. Surprisingly, label smoothing is very good at handling medium ($+2.6\%$ in mAPm) and large ($+9.0\%$ in mAPl) objects. Third, by replacing the Adam to RMSProp~\cite{shi2021rmsprop} optimizer can boost $1.8\%$ in mAP. Interestingly, the benefit of RMSProp largely comes from small and medium objects, but not large objects. For other tested optimizers, i.e., SGD~\cite{ruder2016overview} and AdamW~\cite{loshchilov2018decoupled}, they perform worse than the baseline Adam optimizer. Fourth , several normalization layers (layer-norm~\cite{https://doi.org/10.48550/arxiv.1607.06450}, group-norm~\cite{DBLP:journals/corr/abs-1803-08494}, and instance-norm~\cite{DBLP:journals/corr/UlyanovVL16}) are experimented. We find group-norm (with number of groups $32$) outperforms default batch-norm~\cite{DBLP:journals/corr/IoffeS15} with $0.9\%$ in mAP. Group-norm also performs well on small objects with $1.0\%$ mAPs improvement. But for large objects, batch-norm is still better than others by $2-6\%$ in mAPl. For other normalization layers, they perform either similarly or worse than batch-norm. Lastly, we sweep different number of batch sizes from $8-128$. We observe that batch size equals to $32$ resulting the best mAP with $0.6\%$ improvement. Smaller or larger batch size will degrade the performance. The advantage of increasing the batch size is well-known that it can reduce the training time, enjoy the regularization effects, and reflect more accurate statistics for batch normalization. But using very large batch size remains a challenging problem, especially for object detection~\cite{10.1007/978-3-030-58589-1_29}. Our observation of using very large batch size is consistent with the work in~\cite{DBLP:journals/corr/KeskarMNST16}, where they argued that large batch size may converge to sharper minimizers, lack of explorative properties, and finally lead to performance degradation. We run the same set of experiments for the Faster R-CNN as well. However, these hyperparameters generally lead to worse performance than the baseline except transfer learning leads to improvement compared to the baseline (mAP $+0.5$). This experiment clearly shows one set of optimized hyperparameters for one model may not be optimal for the another.
\begin{table}
\centering
\begin{tabular}{c|cccccc}
\hline
\multirow{2}{*}{Model} & \multicolumn{6}{c}{Metric (\%)} \\ \cline{2-7}
     &   mAP  &  AP$_{50}$     &   AP$_{75}$
     &   mAPs &  mAPm          &   mAPl \\ \hline 
     Baseline & 34.9 & 50.9 & 37.9 & 4.2 & 35.9 & 56.2 \\
     \hline
     Transfer (COCO) & 38.6 (+3.7) & 53.6 & 42.8 & 4.7 & 39.3 & 59.9 \\
     Optimizer (RMSProp) & 36.7 (+1.8) & 52.5 & 40.7 & 5.3 & 37.6 & 55.7 \\
     Smooth ($\alpha=0.2$) & 36.6 (+1.7) & 52.9 & 39.3 & 4.3 & 38.5 & 65.2 \\
     Norm (group-norm) & 35.8 (+0.9) & 52.4 & 39.2 & 5.2 & 35.2 & 53.5 \\
     Batch (size=32) & 35.5 (+0.6) & 51.5 & 39.1 & 4.3 & 36.8 & 61.0 \\
     \hline
\end{tabular}
\caption{Comparison of training a scaled YOLOv4-p5 model with optimized hyperparameters.}
\label{tab:kpi}
\end{table}

\begin{figure}
\centering
\begin{subfigure}[t]{0.49\linewidth}
\includegraphics[clip,keepaspectratio, width=\textwidth]{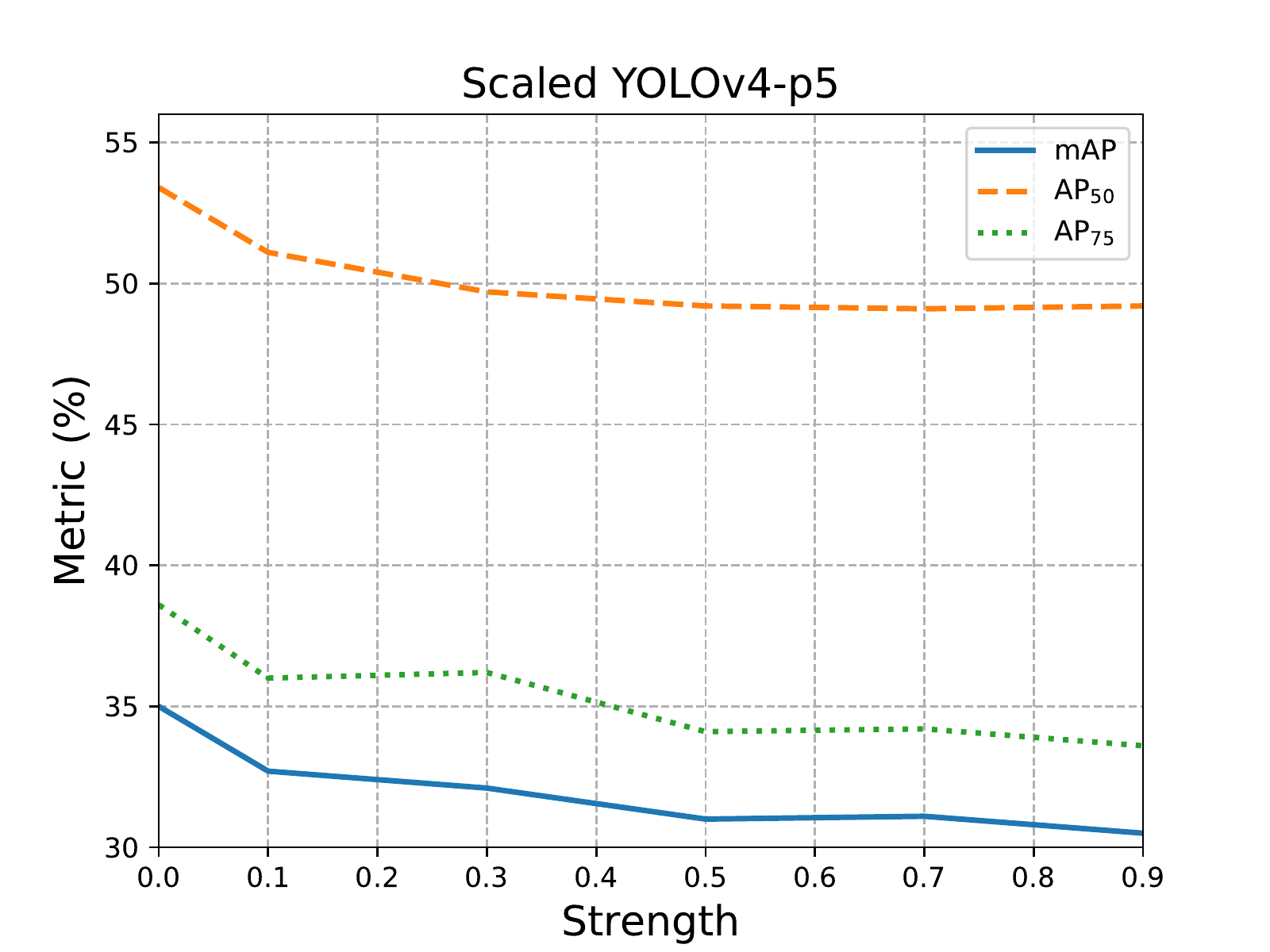}
\caption{Scaled YOLOv4.}
\end{subfigure}
\begin{subfigure}[t]{0.49\linewidth}
\includegraphics[clip,keepaspectratio, width=\textwidth]{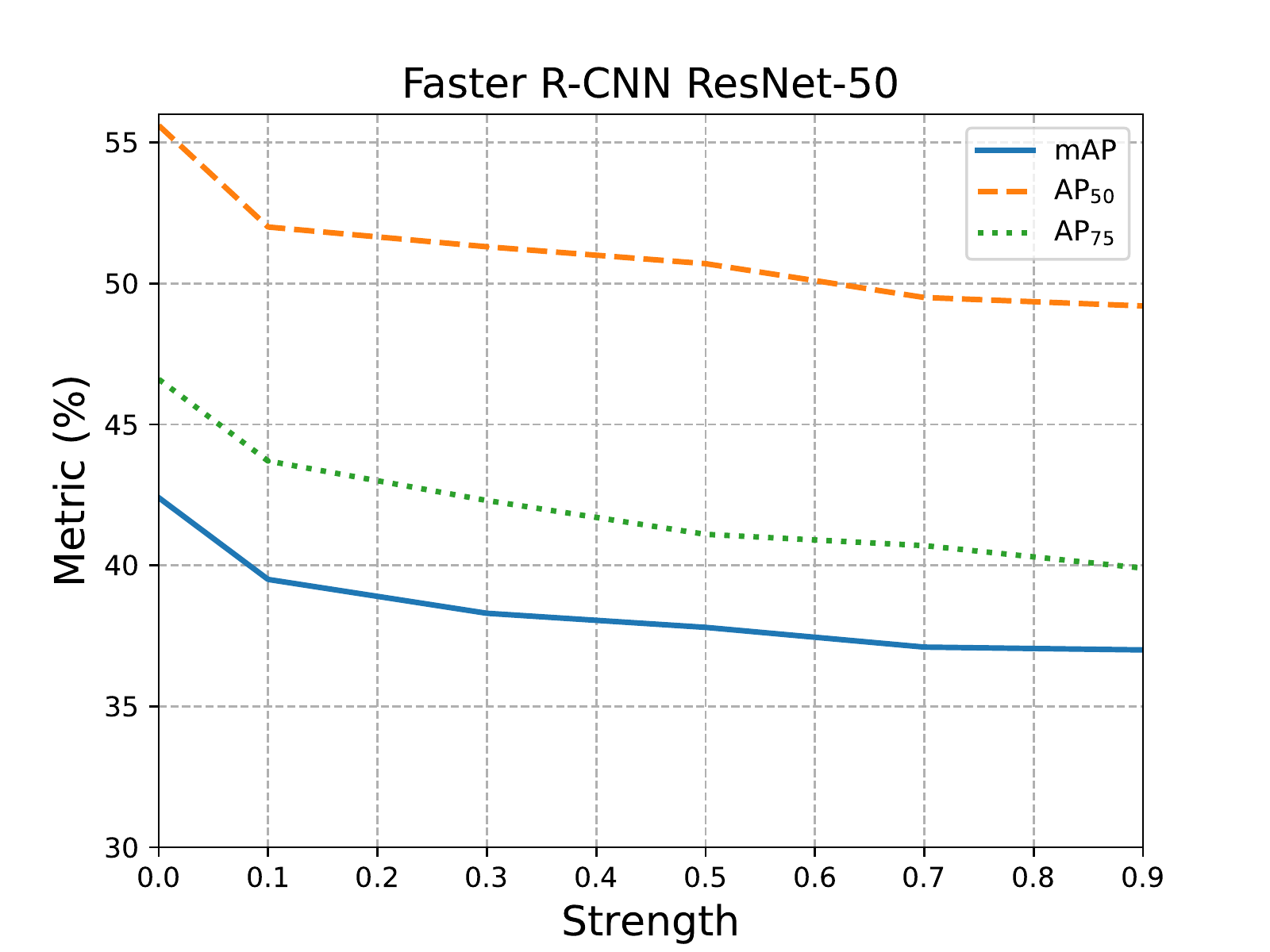}
\caption{Faster R-CNN.}
\end{subfigure}
\caption{Performance metrics versus the strength of color distortion applied on scaled YOLOv4-p5 and Faster R-CNN ResNet-50 models. $0$ strength is equivalent to no color distortion.}
\label{fig:color_distortion}
\end{figure}

\begin{figure}
\begin{subfigure}[t]{0.32\linewidth}
    \includegraphics[width=\linewidth]{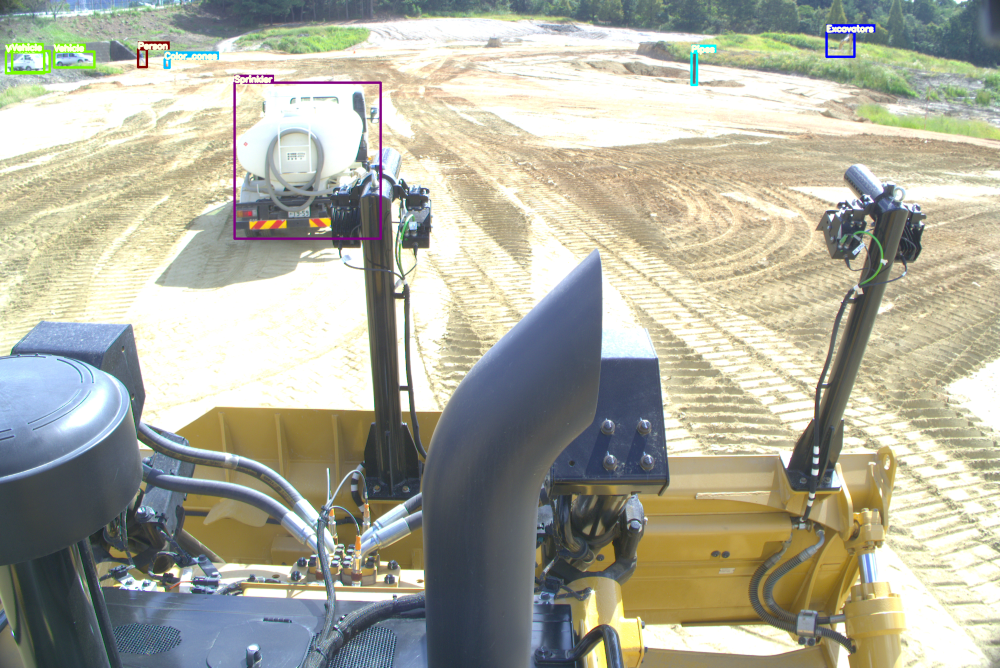}
\label{fig:gt_example_1}
\end{subfigure}\hspace{0.1cm}
\begin{subfigure}[t]{0.32\linewidth}
    \includegraphics[width=\linewidth]{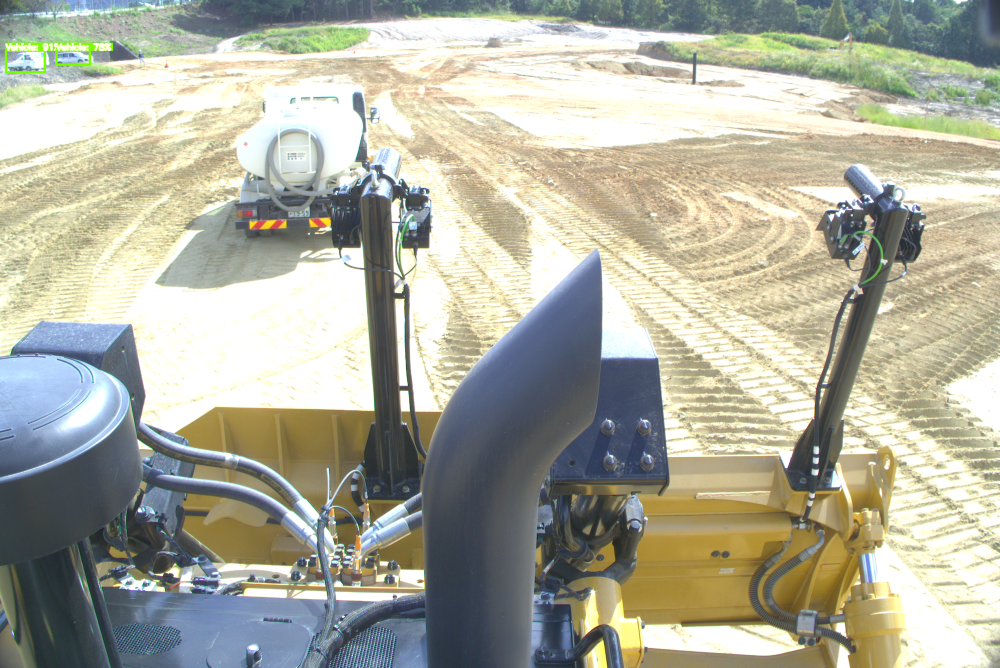}
\label{fig:baseline_example_1}
\end{subfigure}\hspace{0.1cm}
\begin{subfigure}[t]{0.32\linewidth}
    \includegraphics[width=\linewidth]{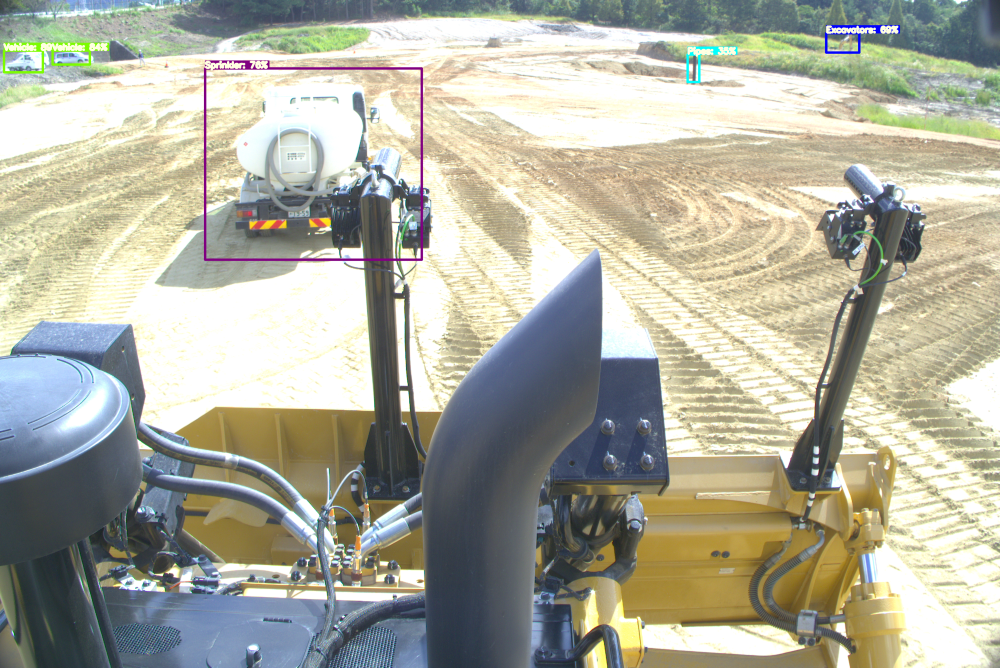}
\label{fig:crop_example_1}
\end{subfigure}\hspace{0.1cm}
\begin{subfigure}[t]{0.32\linewidth}
    \includegraphics[width=\linewidth]{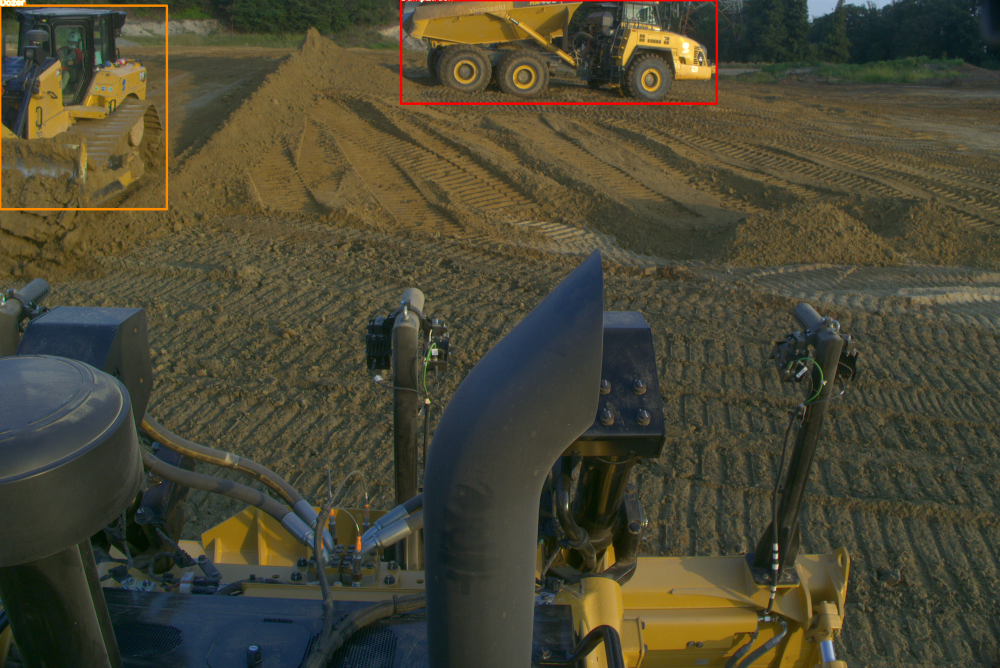}
    \caption{Ground truth.}
\label{fig:gt_example_2}
\end{subfigure}\hspace{0.1cm}
\begin{subfigure}[t]{0.32\linewidth}
    \includegraphics[width=\linewidth]{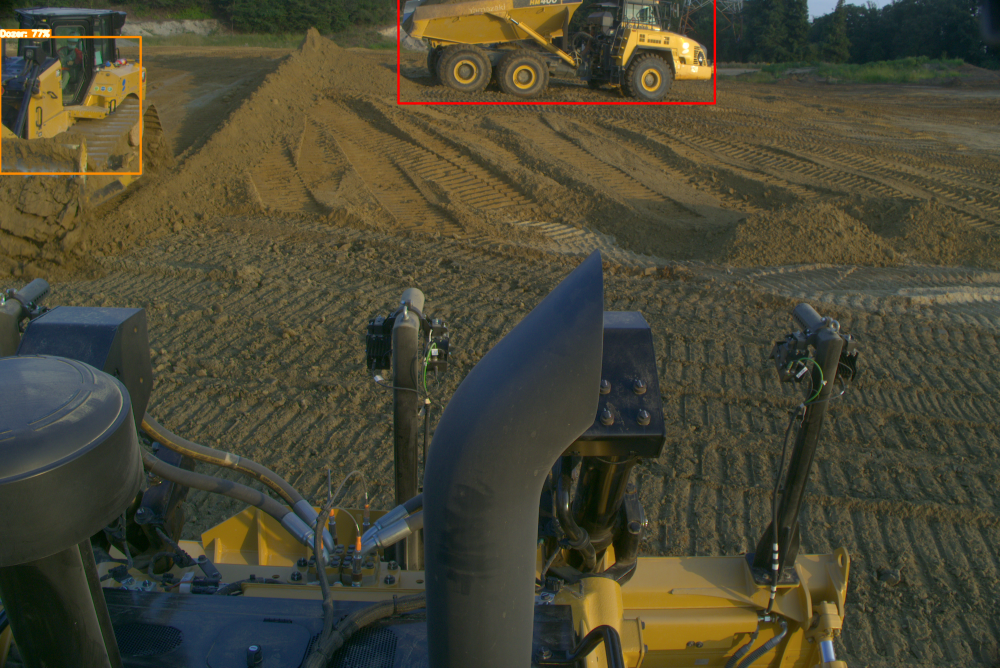}
    \caption{Flip.}
\label{fig:baseline_example_2}
\end{subfigure}\hspace{0.1cm}
\begin{subfigure}[t]{0.32\linewidth}
    \includegraphics[width=\linewidth]{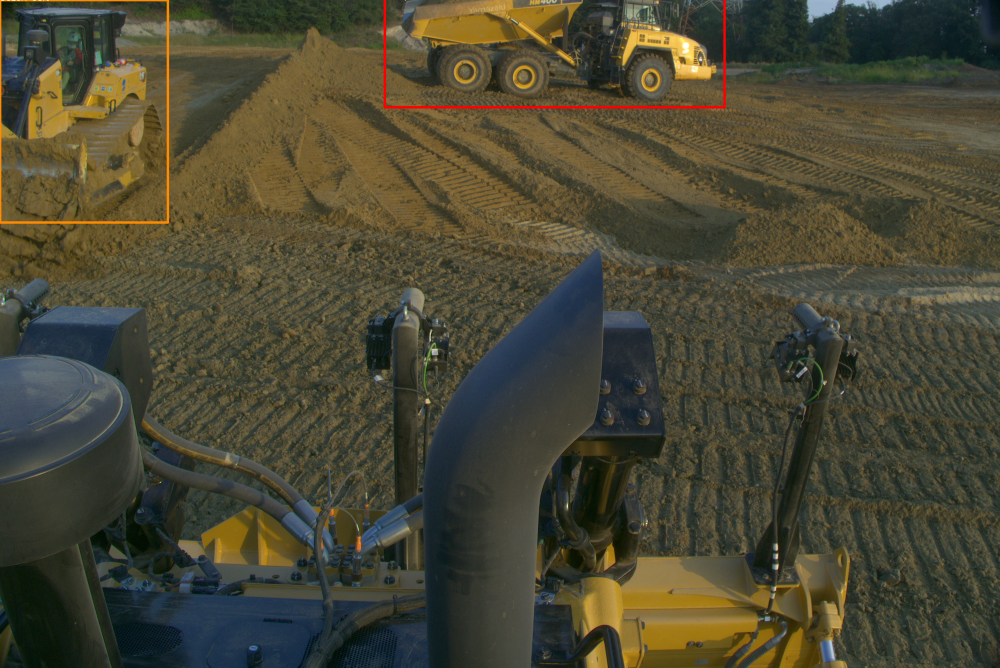}
    \caption{Flip+crop.}
\label{fig:crop_example_2}
\end{subfigure}
\caption{Visual comparisons of augmentations with the scaled YOLOv4-p5 model.
}
\label{fig:augmentations}
\end{figure}

\begin{figure}
\begin{subfigure}[t]{0.32\linewidth}
    \includegraphics[width=\linewidth]{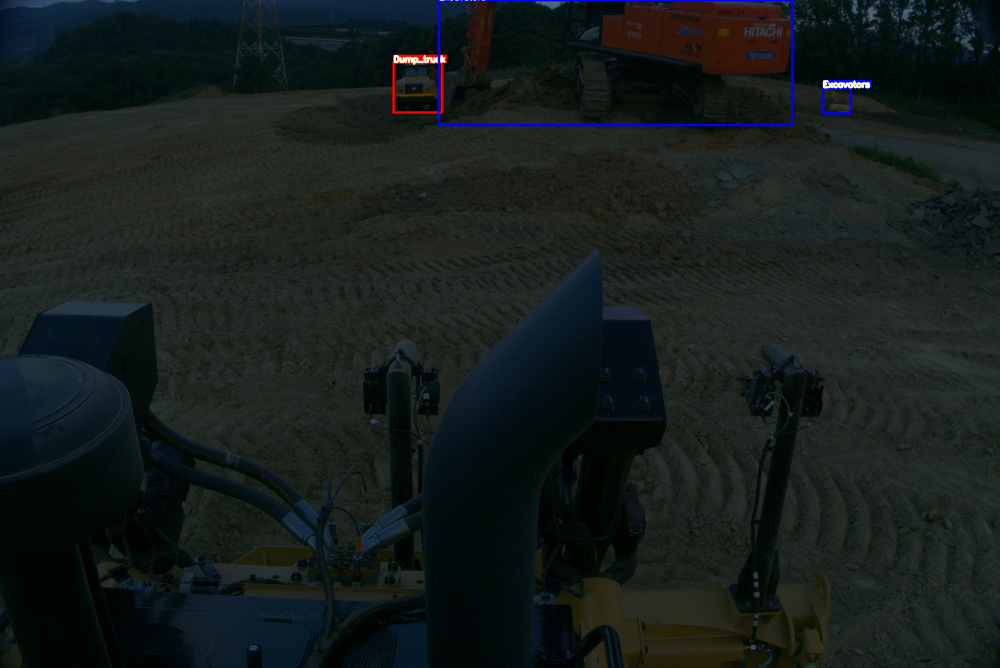}
\label{fig:visual_gt_1}
\end{subfigure}\hspace{0.1cm}
\begin{subfigure}[t]{0.32\linewidth}
    \includegraphics[width=\linewidth]{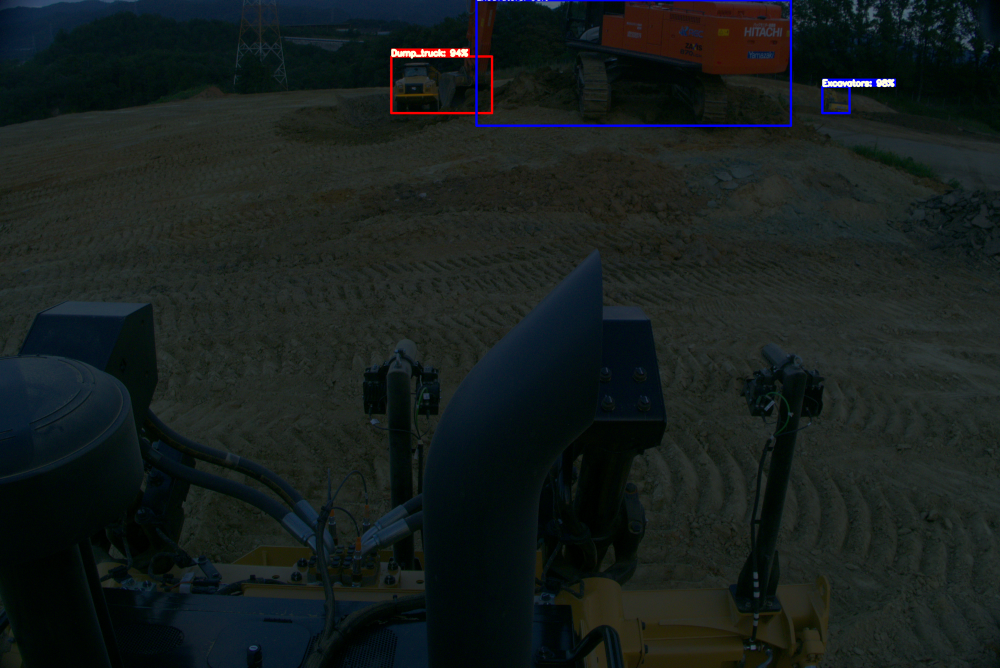}
\label{fig:visual_yolo_1}
\end{subfigure}\hspace{0.1cm}
\begin{subfigure}[t]{0.32\linewidth}
    \includegraphics[width=\linewidth]{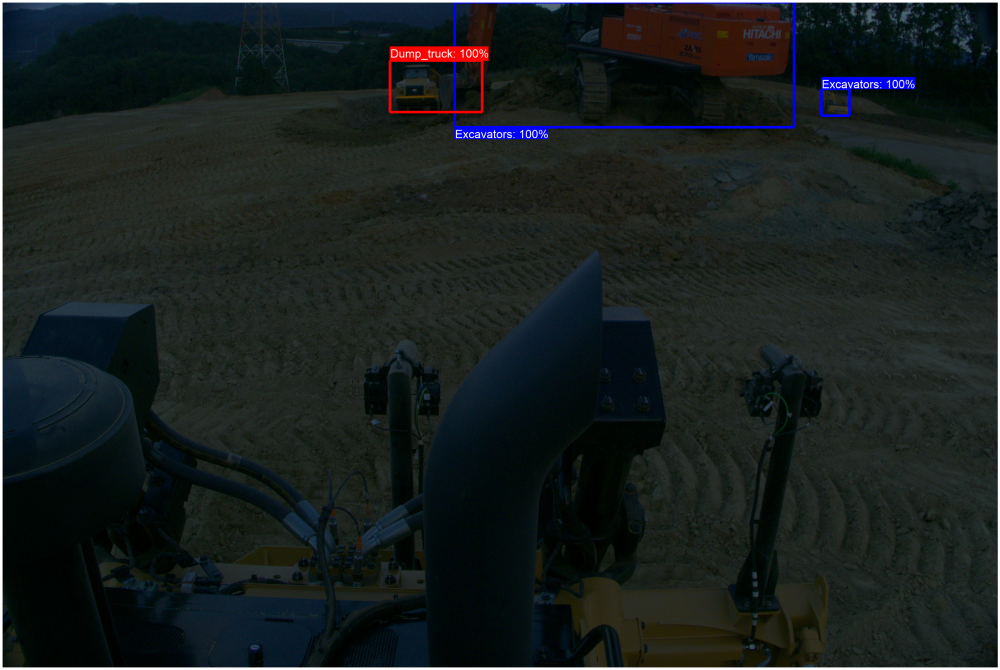}
\label{fig:visual_faster_1}
\end{subfigure}\hspace{0.1cm}
\begin{subfigure}[t]{0.32\linewidth}
    \includegraphics[width=\linewidth]{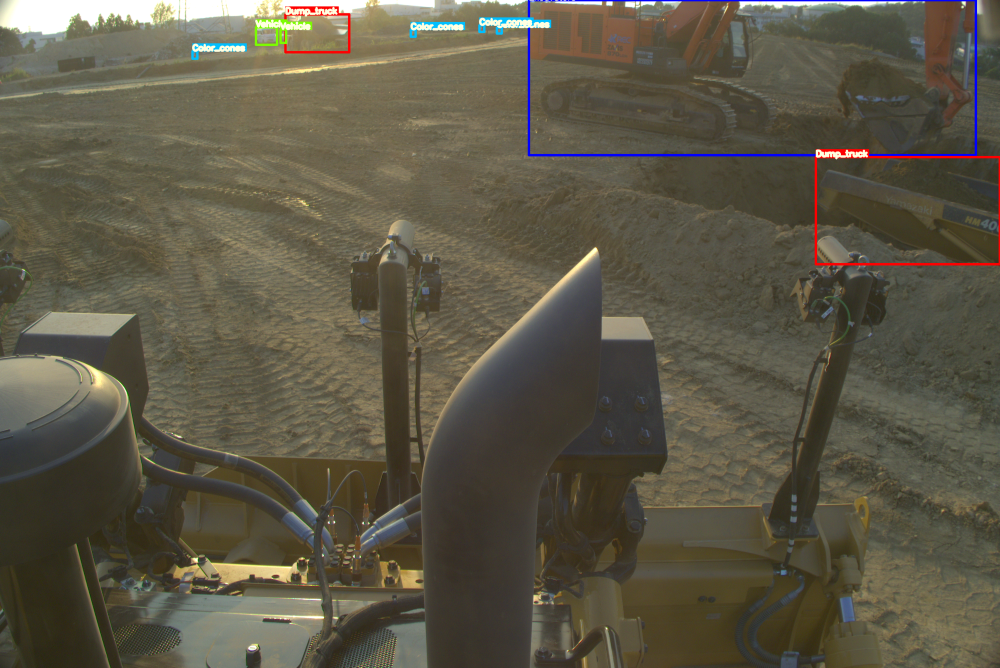}
\label{fig:visual_gt_2}
\end{subfigure}\hspace{0.1cm}
\begin{subfigure}[t]{0.32\linewidth}
    \includegraphics[width=\linewidth]{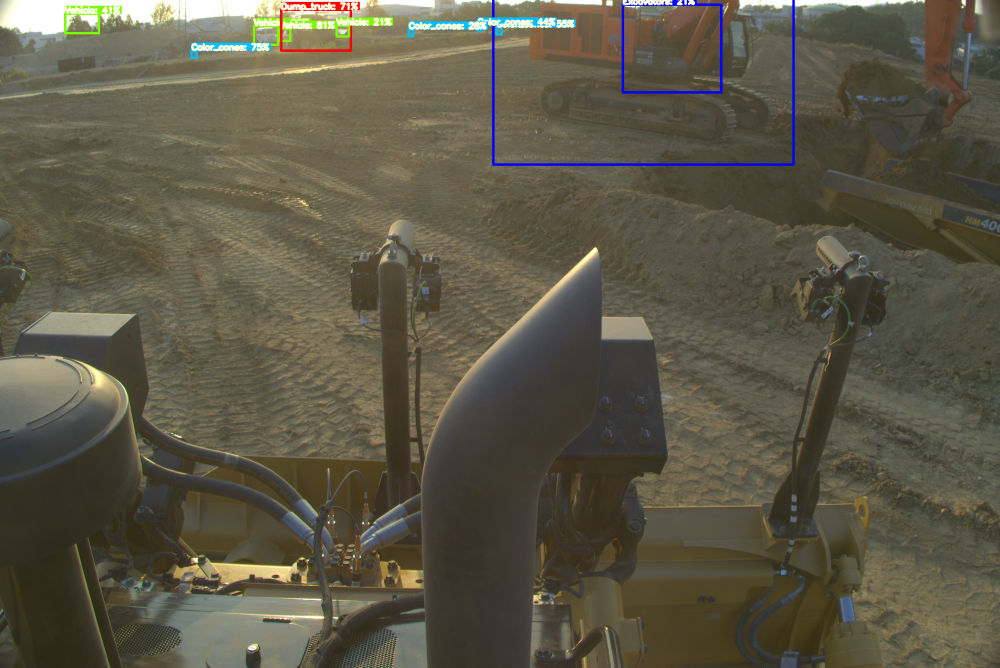}
\label{fig:visual_yolo_2}
\end{subfigure}\hspace{0.1cm}
\begin{subfigure}[t]{0.32\linewidth}
    \includegraphics[width=\linewidth]{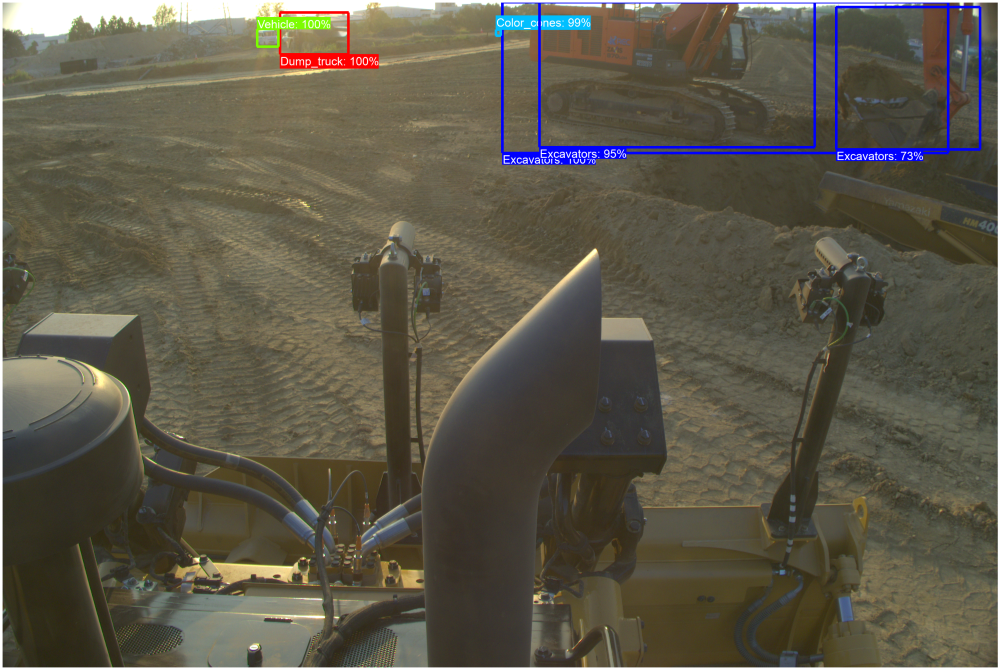}
\label{fig:visual_faster_2}
\end{subfigure}\hspace{0.1cm}
\begin{subfigure}[t]{0.32\linewidth}
    \includegraphics[width=\linewidth]{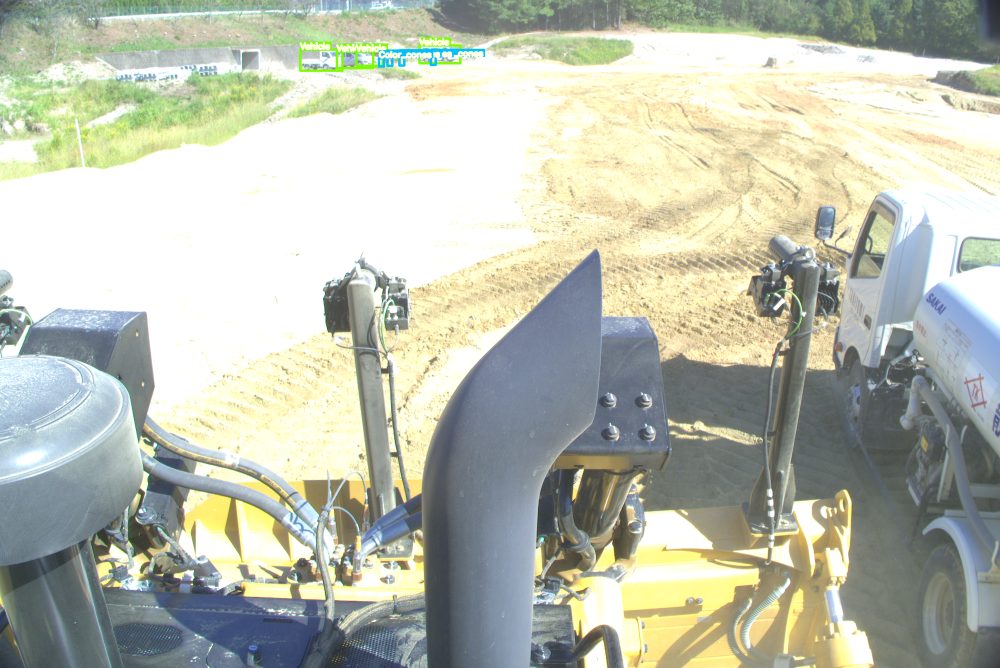}
    \caption{Ground truth.}
\label{fig:visual_gt_3}
\end{subfigure}\hspace{0.1cm}
\begin{subfigure}[t]{0.32\linewidth}
    \includegraphics[width=\linewidth]{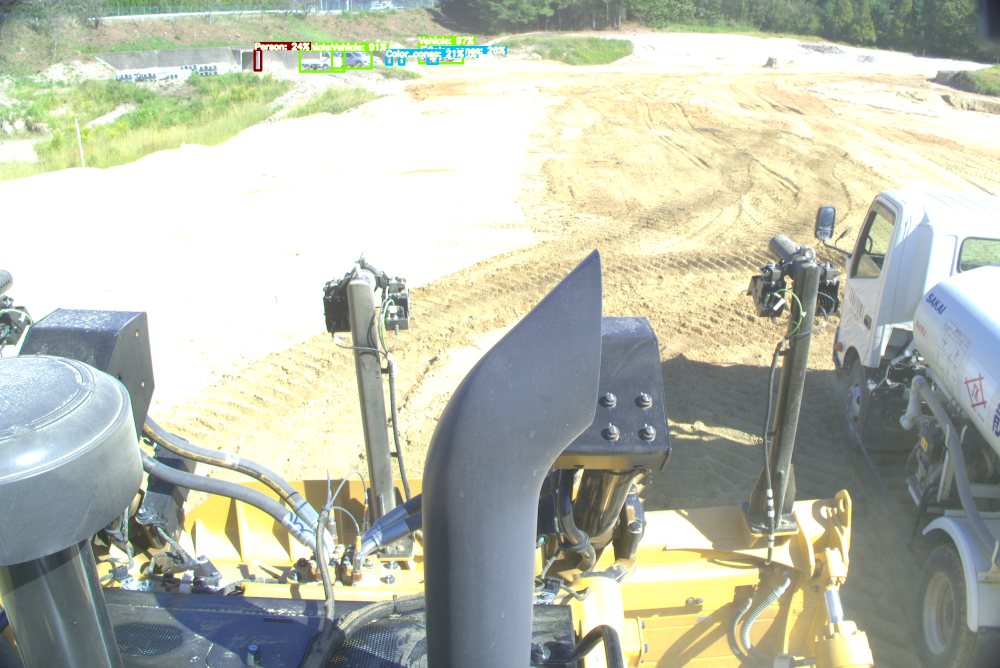}
    \caption{Scaled YOLOv4}
\label{fig:visual_yolo_3}
\end{subfigure}\hspace{0.1cm}
\begin{subfigure}[t]{0.32\linewidth}
    \includegraphics[width=\linewidth]{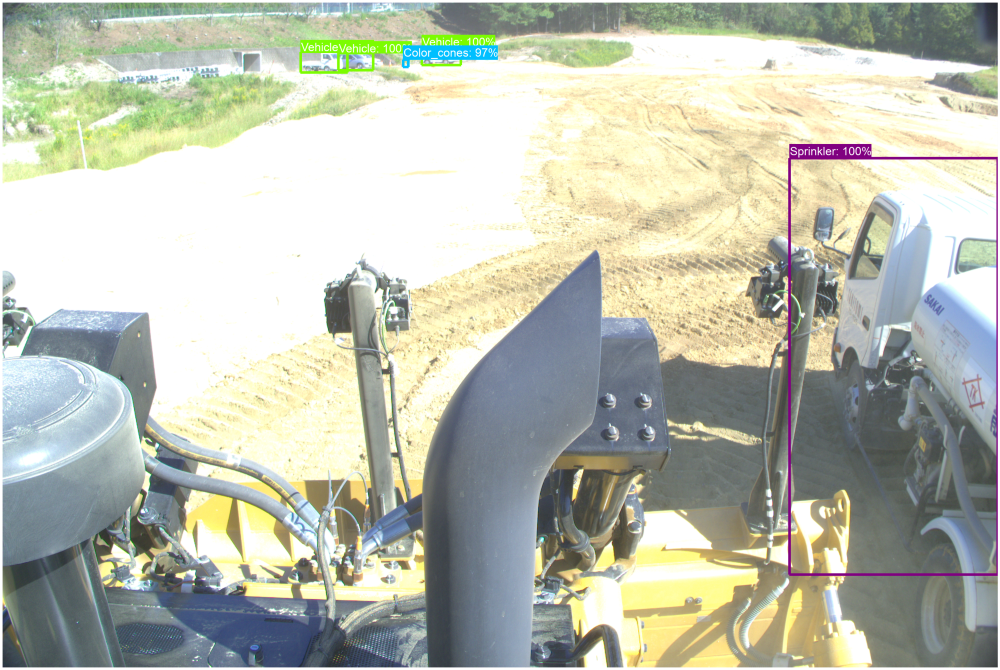}
    \caption{Faster R-CNN}
\label{fig:visual_faster_3}
\end{subfigure}
\caption{Visual examples of baseline models with $1536\times1536$ input size.}
\vspace{-0.5cm}
\label{fig:visual}
\end{figure}

{\bf Augmentations.} We study different augmentation schemes to analyze the best strategy to further improve the performance for the detector. There are two directions for usual image augmentation schemes, namely color distortion and geometric distortion. Color distortion modifies the original image color distribution by changing its brightness, contrast, saturation, hue, etc. The purpose of these color effects is to simulate the rare color distortion scenario that can happen in the dataset such as the strong/low light effect on a construction site. Fig.~\ref{fig:color_distortion} shows the performance with respect to different strength of the color distortion, which we follow the implementation in~\cite{pmlr-v119-chen20j}. It is clear that {\bf color distortion will degrade performance} for both one and two stage models. This may be explained by the fact that our training dataset already captures enough samples for the color distorted scenes. Thus, such additional augmentation is being considered as noise, and it creates out-of-distribution samples that bias the model toward wrong direction. The geometric distortion is done by changing the scaling and shaping of an image, so that different scales and shape of an object can be considered during training. The common geometric distortion includes for example flipping, cropping. shearing, rotation, transformation, etc. We try different combinations of these and the optimized results are presented in Table~\ref{tab:aug_comparison}. Compared to no augmentation, random horizontal flip significantly improves performance for both scaled YOLOv4 and Faster R-CNN. By adding random cropping with cropping scale ratio being chosen from $[0.9, 1.0]$, the model performance for scaled YOLOv4 can be further enhanced. Interestingly, the performance improvement largely comes from mAPl, especially for horizontal flip + crop ($+5.1\%$). This is due to the fact that scaled YOLOv4 model does not work well for large objects that are near the ego-dozer. The reason is we found that the nearby objects are rarely being collected during data collection. Hence it is hard for the model to generalize these types of objects without further processing. With random cropping, the objects can be placed with different scales, and therefore it can alleviate this problem. On the other hand, adding random cropping does not lead to improvement for the Faster R-CNN. Faster R-CNN models are good at detecting large nearby objects (see Fig. \ref{fig:visual_faster_3}). Thus, additional random cropping augmentation does not have similar impacts on Faster R-CNN as it has on scaled YOLOv4. Fig.~\ref{fig:augmentations} presents visual examples to show the effect of using random cropping on scaled YOLOv4. By using horizontal flip only, we can see from the first row the nearby sprinkler can not be detected. From the second row, the bounding box for the nearby dozer does not align with the ground truth position well. However, {\bf by adopting horizontal flip and crop together, we are able to reduce the miss-detection rate for near objects}.

\begin{table}
\centering
\begin{subtable}{\linewidth}
\centering
\begin{tabular}{c|cccccc}
\hline
\multirow{2}{*}{Model} & \multicolumn{6}{c}{Metric (\%)} \\ \cline{2-7}
     &   mAP  &  AP$_{50}$     &   AP$_{75}$
     &   mAPs &  mAPm          &   mAPl \\ \hline 
     No augment & 33.6 & 49.7 & 36.3 & 5.0 & 34.6 & 54.9 \\
     \hline
     Horizontal flip & 34.9 (+1.3) & 50.9 & 37.9 & 4.2 & 35.9 & 56.2 \\
     Horizontal flip+crop & 36.2 (+2.6) & 53.7 & 40.3 & 5.3 & 35.9 & 60.0 \\
     \hline
\end{tabular}
\caption{Scaled YOLOv4-p5.}
\label{tab:yolo_augmentation}
\end{subtable}
\begin{subtable}{\linewidth}
\centering
\begin{tabular}{c|cccccc}
\hline
\multirow{2}{*}{Model} & \multicolumn{6}{c}{Metric (\%)} \\ \cline{2-7}
     &   mAP  &  AP$_{50}$     &   AP$_{75}$
     &   mAPs &  mAPm          &   mAPl \\ \hline 
     No augment & 39.0 & 52.4 & 42.5 & 17.8 & 33.4 & 51.2 \\
     \hline
     Horizontal flip & 42.4 (+3.4)& 55.6 & 46.6 & 18.0 & 36.6 & 62.6\\
     Horizontal flip+crop & 40.2 (+1.2) & 52.9 & 44.1 & 17.4 & 34.8 & 54.3 \\
     \hline
\end{tabular}
\caption{Faster R-CNN ResNet-50.}
\label{tab:faster_rcnn_augmentation}
\end{subtable}
\caption{Comparison of various augmentation schemes.}
\label{tab:aug_comparison}
\end{table}

\begin{figure}
\begin{subfigure}[t]{0.32\linewidth}
    \includegraphics[width=\linewidth]{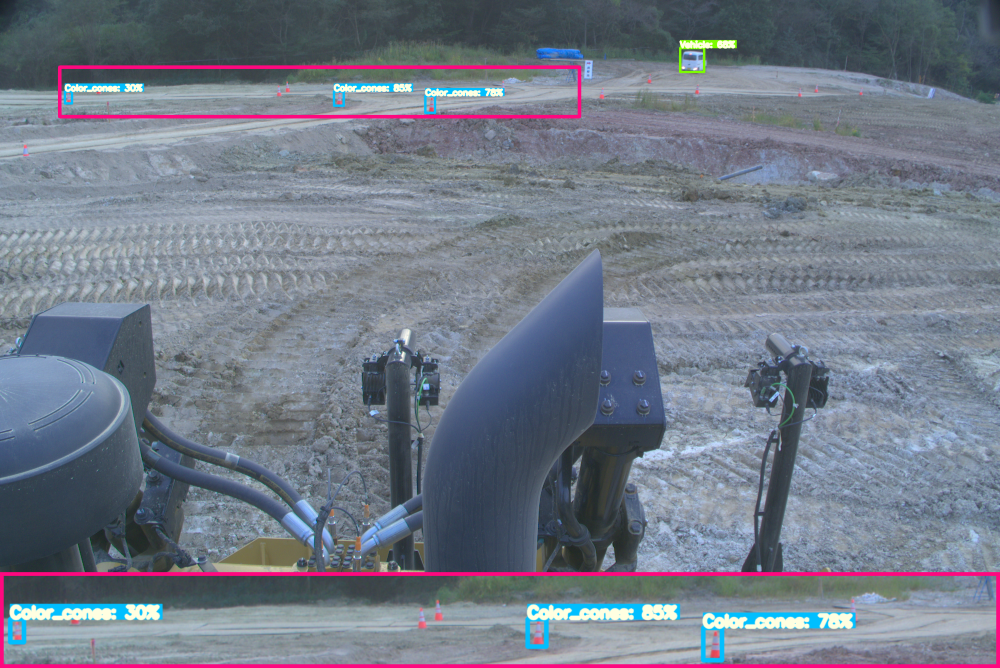}
    \caption{$416\times416$.}
\label{fig:416}
\end{subfigure}\hspace{0.1cm}
\begin{subfigure}[t]{0.32\linewidth}
    \includegraphics[width=\linewidth]{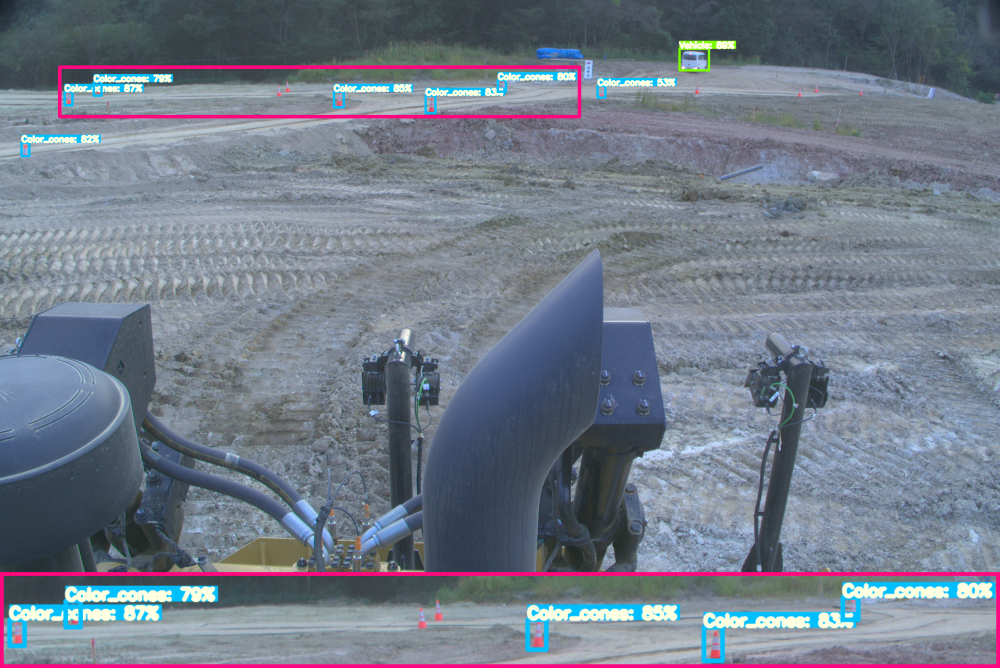}
    \caption{$512\times512$.}
\label{fig:512}
\end{subfigure}\hspace{0.1cm}
\begin{subfigure}[t]{0.32\linewidth}
    \includegraphics[width=\linewidth]{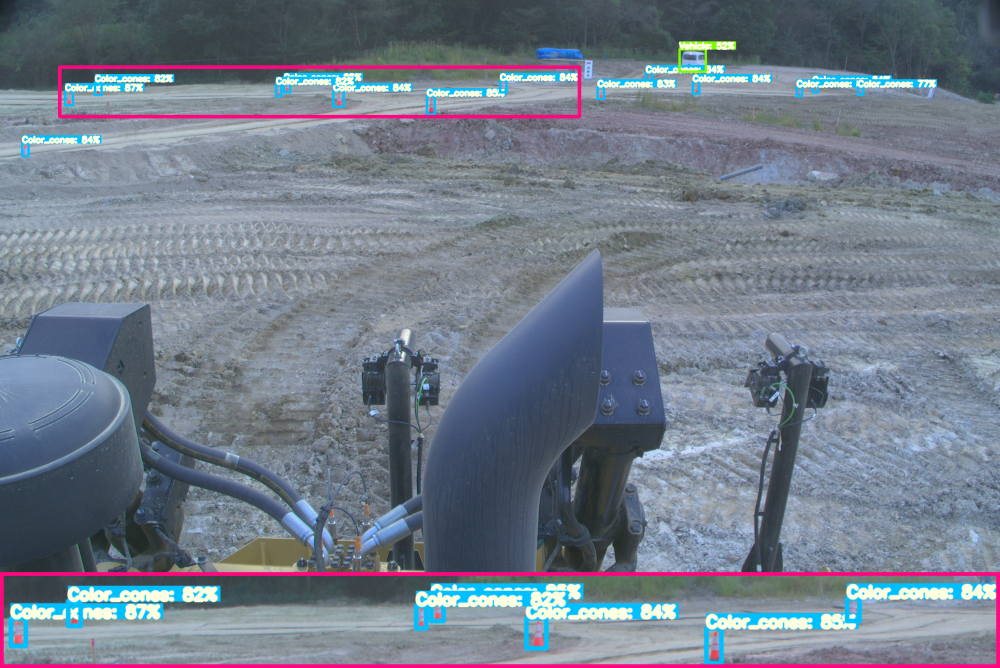}
    \caption{$768\times768$.}
\label{fig:768}
\end{subfigure}
\caption{Detection with scaled YOLOv4-p5 model on various input resolutions. The region inside the pink bounding box is zoomed in for better visualization.}
\label{fig:input_resolution}
\end{figure}

\begin{figure}
\centering
\begin{subfigure}[t]{0.49\linewidth}
\includegraphics[clip,keepaspectratio, width=\textwidth]{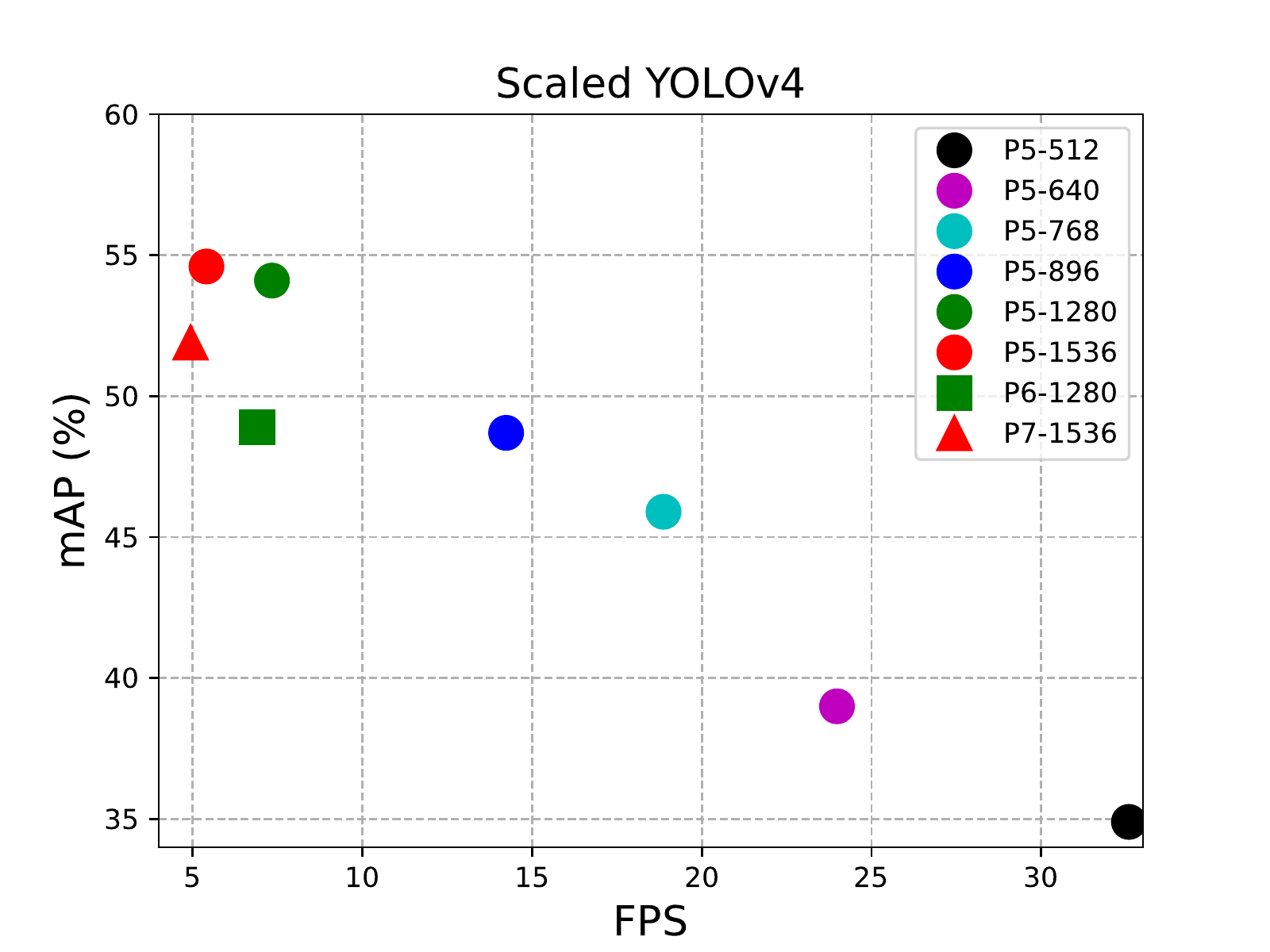}
\caption{Scaled YOLOv4.}
\end{subfigure}
\begin{subfigure}[t]{0.49\linewidth}
\includegraphics[clip,keepaspectratio, width=\textwidth]{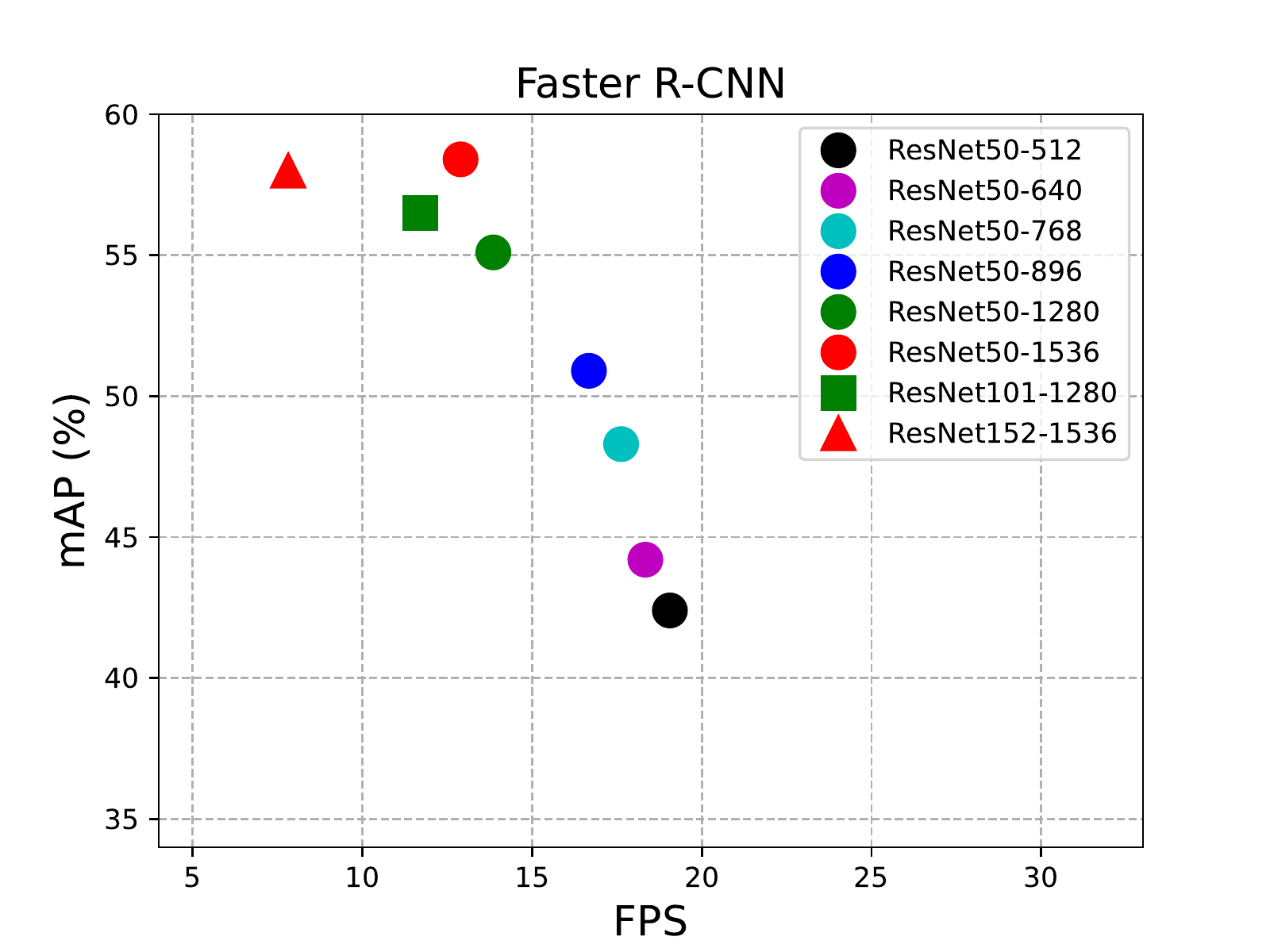}
\caption{Faster R-CNN.}
\end{subfigure}
\caption{Performance comparisons for model accuracy versus inference speed with different model and input scales. Inference speed is measured with FPS.}
\label{fig:model_scaling}
\end{figure}

{\bf Input Scaling.} Table~\ref{tab:one_stage_resolution} and~\ref{tab:two_stage_resolution} show the impact of input resolution on detection accuracy for scaled YOLOv4 and Faster R-CNN models, respectively. It can be observed that {\bf input resolution is the most important factor regarding detection performance.} The improvement is more obvious for small and medium objects. This is expected since smaller objects need higher resolution feature maps to capture enough information. On the other hand, models with higher input resolutions suffer more on inference latency (see Fig. \ref{fig:model_scaling} for latency reference). Note we do not try beyond input size $1536\times1536$, since this is the largest scale that the original scaled-YOLOv4 used for their model. Fig.~\ref{fig:input_resolution} shows detection results for scaled YOLOv4 using a representative image from the validation set. We found there are many images on construction sites showing lots of color-cones, and color-cones are the tiniest objects among all classes. From the figures it is obvious that if we increase the input image resolution, we can detect more color-cones. To improve the detection for small objects is the key to increasing the overall performance, and the visualization shows consistent conclusion as the quantitative results in Table~\ref{tab:one_stage_resolution} and~\ref{tab:two_stage_resolution}.

\begin{table}
\centering
\begin{subtable}{\linewidth}
\centering
\begin{tabular}{c|cccccc}
\hline
\multirow{2}{*}{Model} & \multicolumn{6}{c}{Metric (\%)} \\ \cline{2-7}
     &   mAP  &  AP$_{50}$     &   AP$_{75}$
     &   mAPs &  mAPm          &   mAPl \\ \hline 
     $416\times416$ & 30.4 & 44.3 & 33.2 & 1.0 & 30.3 & 51.5 \\
     \hline
     $512\times512$ & 34.9 & 50.9 & 37.9 & 4.2 & 35.9 & 56.2 \\
     \hline
     $640\times640$ & 39.0 & 58.3 & 42.2 & 11.8 & 39.9 & 56.2 \\
     \hline
     $768\times768$ & 45.9 & 68.2 & 50.0 & 22.9 & 42.4 & 56.4 \\
     \hline
     $896\times896$ & 48.7 & 70.0 & 54.6 & 25.6 & 46.0 & 59.0 \\
     \hline
     $1280\times1280$ & 54.1 & 73.8 & 60.3 & 29.4 & 53.7 & {\bf 63.5} \\
     \hline
     $1536\times1536$ & {\bf 54.6} & {\bf 76.8} & {\bf 61.3} & {\bf 32.3} & {\bf 53.8} & 62.8 \\
     \hline
\end{tabular}
\caption{Scaled YOLOv4-p5.}
\label{tab:one_stage_resolution}
\end{subtable}
\begin{subtable}{\linewidth}
\centering
\begin{tabular}{c|cccccc}
\hline
\multirow{2}{*}{Model} & \multicolumn{6}{c}{Metric (\%)} \\ \cline{2-7}
     &   mAP  &  AP$_{50}$     &   AP$_{75}$
     &   mAPs &  mAPm          &   mAPl \\ \hline 
     $416\times416$ & 34.6 & 47.1 & 37.1 & 14.3 & 28.9 & 49.8 \\
     \hline
     $512\times512$ & 42.4& 55.6 & 46.6 & 18.0 & 36.6 & 62.6 \\
     \hline
     $640\times640$ & 44.2 & 59.2 & 48.2 & 26.6 & 40.2 & 66.6 \\
     \hline
     $768\times768$ & 48.3 & 63.8 & 53.3 & 25.3 & 45.8 & 62.8 \\
     \hline
     $896\times896$ & 50.9 & 67.3 & 56.2 & 31.0 & 47.6 & 63.2 \\
     \hline
     $1280\times1280$ & 55.1 & 71.9 & 60.9 & 31.5 & 51.6 &  70.1 \\
     \hline
     $1536\times1536$ & {\bf 58.4} & {\bf 75.5} & {\bf 64.2} & {\bf 34.5} & {\bf 54.2} & {\bf 72.4} \\
     \hline
\end{tabular}
\caption{Faster R-CNN ResNet-50.}
\label{tab:two_stage_resolution}
\end{subtable}
\caption{Comparison of different input resolutions. {\bf Bold} text represents the best score.}
\vspace{-1cm}
\label{tab:input_resolution}
\end{table}
{\bf Model scaling.} The detection accuracy versus inference speed is presented in Fig.~\ref{fig:model_scaling}. The inference speed is measured on one NVIDIA Tesla V100 GPU, and the inference duration includes both the model and post-processing (Non-Maximum Suppression) time. Note scaled-YOLOv4 has several versions, i.e., 3 heads (P5), 4 heads (P6), and 5 heads (P7) for detection. Larger model adopts larger input resolution according to the original paper~\cite{Wang_2021_CVPR}, e.g., P5 is with $896\times896$, P6 is with $1280\times1280$, and P7 is with $1536\times1536$. Similarly, Faster R-CNN has different backbone models, i.e., ResNet-50, ResNet-101, and ResNet-152. For scaled YOLOv4, we can observe that mAP only increases slightly from P5-896 to P6, but the speed decreases by half. However, there is a $3\%$ increase in mAP from P6 to P7. We further run the experiments to see the scaling for P5 itself, and we find that {\bf scaling P5 up to resolutions that are the same as P6 and P7 results in better mAP and speed}. The mAP improvement is $5.2\%$ in P6 and $2.7\%$ in P7. This may be explained by that using more heads does not help detection in our dataset. The reason is that deeper heads are good at capturing large objects but not small objects. Since our challenge is in small objects, more complex model can overfit to the training data easily. For Faster R-CNN, ResNet-50 with different resolutions shows an expected mAP versus frame per second (FPS) behavior. With increasing resolution, we have higher mAP and lower FPS. More importantly, ResNet-50 with $1536\times 1536$ outperforms counterpart ResNet-101 with $1280\times1280$ and ResNet-152 with $1536\times 1536$ in terms of both mAP and FPS.

{\bf Visualization.}
We present three visual examples to compare scaled YOLOv4 and Faster R-CNN models in Fig.~\ref{fig:visual}. Each row represents one example. The first row is an example under dark light scenario, i.e., when the dozer works during sunset period. We can observe both models can capture dump-trucks and excavators correctly, which shows the robustness to light variations. The second row example demonstrates a large and segmented excavator, and it is out of the image border. We can infer its head and body should be considered as one object by human easily, however, scaled YOLOv4 model can only recognize its body. On the other hand, Faster R-CNN can capture the whole detection. Although both body and head are detected as redundant objects, it is capable of connecting two parts and joint them as a single object. The third row shows an example of a very near object coming across the ego-vehicle. We notice that the ground truth image misses the label for the nearby sprinkler by random labeling error, and scaled YOLOv4 can not detect such object. Nevertheless, Faster R-CNN can identify this kind of closer object to ego-vehicle. We speculate that, unlike scaled YOLOv4, Faster R-CNN trains to generate more flexible and reliable proposals by its first stage RPN, and it can better capture very large and small objects which does not fully rely on pre-defined anchor box sizes. This may be the main reason that we see both mAPs and mAPl for Faster R-CNN are better than scaled YOLOv4. Especially for mAPl it can add $9.6\%$ improvement under $1536\times1536$ input resolution.

\section{Conclusions}
We introduce a new type of AV for construction work. The sensor setup, data collection, and labeling strategies are presented. Detailed data statistics are also revealed to understand labels thoroughly. Two commonly adopted OD models are compared in terms of detection performance, and several key performance factors such as transfer learning, data augmentation, and input/model scaling, are highlighted by running extensive experiments. We hope this work provides guidelines and insights for setting up an autonomous construction vehicle for perception modules, and it can also serve as a baseline in terms of object detection performance.

For the future work we plan to address the issue of highly imbalanced dataset and how to efficiently exploit unlabeled video frames taken from the camera. Furthermore, we would like to study multiple camera/LiDAR fusion to create a single robust detection model. Finally, we would like to investigate the feasibility of domain adaptation from different construction sites, to ensure the models can generalize in different use cases.

%
%
\bibliographystyle{splncs04}
\bibliography{egbib}
\end{document}